\documentclass[letterpaper]{article} 
\usepackage{aaai2026}  
\usepackage{times}  
\usepackage{helvet}  
\usepackage{courier}  
\usepackage[hyphens]{url}  
\usepackage{graphicx} 
\usepackage{natbib}  
\usepackage{caption} 
\usepackage{algorithm}
\usepackage{algorithmic}

\usepackage{newfloat}
\usepackage{listings}
\DeclareCaptionStyle{ruled}{labelfont=normalfont,labelsep=colon,strut=off} 
\floatstyle{ruled}
\newfloat{listing}{tb}{lst}{}
\floatname{listing}{Listing}
\usepackage{enumitem} 
\usepackage{xspace} 
\usepackage{bm} 
\usepackage{amsthm} 
\usepackage{amssymb} 
\usepackage{mathtools} 
\usepackage{makecell} 
\usepackage{booktabs} 
\usepackage{subcaption} 

\title{Multi-Agent Pointer Transformer: Seq-to-Seq Reinforcement Learning for Multi-Vehicle Dynamic Pickup-Delivery Problems}
\author{
    Zengyu Zou\textsuperscript{\rm 1, \rm 4}, Jingyuan Wang\textsuperscript{\rm 1, \rm 2, \rm 3, \rm 4, \thanks{Corresponding author}}, Yixuan Huang\textsuperscript{\rm 1, \rm 4}, Junjie Wu\textsuperscript{\rm 2, \rm 3}
}
\affiliations{
    \textsuperscript{\rm 1}School of Computer Science and Engineering, Beihang University, Beijing, China \\
    \textsuperscript{\rm 2}School of Economics and Management, Beihang University, Beijing, China \\
    \textsuperscript{\rm 3}MIIT Key Laboratory of Data Intelligence and Management, Beihang University, Beijing, China \\
    \textsuperscript{\rm 4}MOE Engineering Research Center of Advanced Computer Application Technology, Beihang University, China \\
    
    \{zzy0001213, jywang, yixuanhuang, wujj\}@buaa.edu.cn
}

\usepackage{bibentry}

\begin{document}

\maketitle

\newcommand{\problem}{MVDPDPSR\xspace}
\newcommand{\ie}{i.e.\xspace} 
\theoremstyle{definition} 
\newtheorem{definition}{Definition}

\begin{abstract}
This paper addresses the cooperative Multi-Vehicle Dynamic Pickup and Delivery Problem with Stochastic Requests (MVDPDPSR) and proposes an end-to-end centralized decision-making framework based on sequence-to-sequence, named Multi-Agent Pointer Transformer (MAPT).
MVDPDPSR is an extension of the vehicle routing problem and a spatio-temporal system optimization problem, widely applied in scenarios such as on-demand delivery.
Classical operations research methods face bottlenecks in computational complexity and time efficiency when handling large-scale dynamic problems.
Although existing reinforcement learning methods have achieved some progress, they still encounter several challenges: 1) Independent decoding across multiple vehicles fails to model joint action distributions; 2) The feature extraction network struggles to capture inter-entity relationships; 3) The joint action space is exponentially large.
To address these issues, we designed the MAPT framework, which employs a Transformer Encoder to extract entity representations, combines a Transformer Decoder with a Pointer Network to generate joint action sequences in an AutoRegressive manner, and introduces a Relation-Aware Attention module to capture inter-entity relationships.
Additionally, we guide the model's decision-making using informative priors to facilitate effective exploration. Experiments on 8 datasets demonstrate that MAPT significantly outperforms existing baseline methods in terms of performance and exhibits substantial computational time advantages compared to classical operations research methods.
\end{abstract}

\begin{links}
    \link{Code}{https://github.com/Beihang-BIGSCity/MAPT}
\end{links}

\section{Introduction}

The Pickup and Delivery Problem is a type of vehicle routing problem that has demonstrated its importance in many real-world applications, such as online food delivery.
In real-world scenarios, the arrival time, origin, and destination of requests are often unpredictable, and there are strict requirements for service response times.
Therefore, efficiently planning vehicle routes based on real-time updated information is particularly important in solving this problem.
This work focuses on solving the cooperative Multi-Vehicle Dynamic Pickup and Delivery Problem with Stochastic Requests (\problem). Unlike the traditional Pickup and Delivery Problem, in this problem, we need to route multiple vehicles, and the occurrence times of all requests are unknown in advance. Moreover, the origin and destination of each request are only revealed after the request emerges.

Existing methods for addressing dynamic vehicle routing problems fall into three categories: \textbf{heuristic methods}, \textbf{classical operations research methods}, and \textbf{reinforcement learning-based algorithms}.
\textbf{Heuristic methods} have shown some success \cite{sheridan2013dynamic, fikar2018decision, andersson2021comparative} by routing vehicles through manually designed heuristic rules, such as the nearest-distance rule. Their strengths are simplicity and efficiency, fitting real-time decision scenarios, but they have clear limits as they depend heavily on manually designed rules and struggle to ensure optimality.
\textbf{Classical operations research methods} typically adopt the Rolling-Horizon paradigm, which transforms dynamic problems into static ones. These static subproblems are then solved using either exact algorithms \cite{lu2004exact, liu2018dynamic, savelsbergh1998drive} or metaheuristics \cite{schilde2011metaheuristics, geiser2020best, cai2022efficient}.
This paradigm requires accumulating a sufficient number of requests before initiating planning, which fails to meet the needs of real-time dynamic decision-making. Also, both exact and metaheuristic algorithms have high computational complexity, greatly affecting their time efficiency in practice.
With technology development, \textbf{reinforcement learning-based algorithms} have been used for multi-vehicle routing. Multi-agent reinforcement learning methods have been proposed for the static multi-vehicle pickup and delivery problem \cite{zong2022mapdp, berto2024parco}. They encode the current state with an Encoder, decode independently for each vehicle with a Decoder, and use a conflict handler in post-processing to solve vehicle conflicts.

However, these multi-agent methods still face the following issues:
(1) Existing multi-agent reinforcement learning methods decode each agent independently, failing to model the joint probability distribution of agents' actions. Furthermore, the decisions generated by these methods may conflict among agents (e.g., competing for the same request).
(2) Existing feature extraction networks fail to capture relationships between entities (e.g., stations, vehicles, requests), which are key components of the \problem.  
(3) The joint action space is often exponentially large, which prevents reinforcement learning from exploring useful actions.

To tackle these remaining issues, we propose the Multi-Agent Pointer Transformer (MAPT), an end-to-end centralized decision-making framework based on sequence-to-sequence to solve \problem.
Specifically, we model the \problem problem as a Markov Decision Process, use a Transformer Encoder to extract entity representations, and then combine a Transformer Decoder with a Pointer Network to convert the embedding sequence into a joint action sequence in an AutoRegressive manner.
To extract relationships between entities, we propose a Relation-Aware Attention module that embeds entity relationships and adds them to the attention matrix.
To address the challenge of reinforcement learning exploring a vast joint action space, we design informative priors based on load balancing and station distance, and fuse these priors with the action distribution decoded by the Decoder, enabling the model to select actions with higher potential rewards. 

In summary, the main contributions of our work are as follows:
\begin{itemize}[leftmargin=*]
  \item We formalize the \problem problem as a Markov Decision Process and design the Multi-Agent Pointer Transformer (MAPT) framework to decode joint actions and model the joint probability.
  \item We design a Relation-Aware Attention module to capture various relationships between entities. 
  \item We design informative priors and integrate them into the model's decision distribution to facilitate effective exploration.
  \item We validate the effectiveness of MAPT on 8 datasets, and experiments show that MAPT outperforms various baselines and demonstrates significant computational time advantages over classical operations research methods.
\end{itemize}

\section{Preliminaries}
\subsection{Problem Formulation}

In this section, we formally define the Multi-Vehicle Dynamic Pickup and Delivery Problem with Stochastic Requests(\problem).
A typical \problem scenario involves a fleet of vehicles (Multi-Vehicle) tasked with picking up and delivering cargos among a set of stations.
The pickup and delivery requests arise dynamically over time, reflecting the ``Dynamic'' nature of the problem. Accordingly, stations, vehicles, and requests constitute the three fundamental components of a \problem. We provide their formal definitions below.

\begin{definition}[Stations]
    The stations in \problem are represented as a fully connected weighted graph $\mathcal{G} = \{N, \bm{E}\}$. Here, $N = \{n_1,$ $\ldots,$ $n_i,$ $\ldots,$ $n_I\}$ denotes the set of stations, where $n_i$ represents the $i$-th station. $\bm{E} \in \mathbb{R}^{|N| \times |N|}$ is the distance matrix, where each element $e_{n_i, n_j}$ indicates the travel time from station $n_i$ to station $n_j$. 
    In the station graph $\mathcal{G}$, each station can serve as either an origin (pickup station) or a destination (delivery station) for cargos.
\end{definition}

\begin{definition}[Vehicles]
    A \problem scenario involves $K$ vehicles, where each vehicle $v_k$ is represented by a tuple of four state variables:
    \begin{equation}
    v_k = \left\langle\; {cap}_k,\; {spa}_k,\; {to}_k,\; {dist}_k\; \right\rangle.
    \end{equation}
    The meanings of these variables are as follows:
    \begin{itemize}[leftmargin=*]
        \item ${cap}_k$ (capacity): the total capacity of vehicle $v_k$.
        \item ${spa}^t_k$ (space): the remaining available space of vehicle $v_k$ at time \(t\). For notational brevity, we omit the superscript time index \(t\) when unambiguous (e.g. \(spa_k\)).
        \item ${to}_k$ (destination): the current destination station of the vehicle.
        \item ${dist}_k$ (distance): the remaining distance (travel time) to reach the current destination.
    \end{itemize}
    If ${dist}_k \neq 0$, the vehicle is en route; otherwise, it is located at a station. Notably, the destination of a vehicle cannot be changed while it is en route.
\end{definition}

\begin{definition}[Requests]
    The objective of \problem is to fulfill $M$ cargo delivery requests, where each request $r_m$ is represented by a tuple of six state variables:
    \begin{equation}
        r_m = \left\langle \; {from}_m,\; {to}_m,\; {val}_m,\; {vol}_m,\; {time}_m, \; {state}_m \;\right\rangle.
    \end{equation}
    The meanings of these variables are as follows:
    \begin{itemize}[leftmargin=*]
    \item ${from}_m$ (origin): the origin station of request $r_m$.
    \item ${to}_m$ (destination): the destination station of request $r_m$.
    \item ${val}_m$ (value): the profit associated with fulfilling the request.
    \item ${vol}_m$ (volume): the cargo volume required for the request, which consumes vehicle space.
    \item ${time}_m$ (appearance time): the time at which the request becomes visible to the system.
    \item ${state}_m$ (state): the current state of request $r_m$.
    \end{itemize}
    A request becomes visible only when the current time reaches its appearance time ${time}_m$. The state variable ${state}_m \in \{\mathrm{unassigned}, \mathrm{picked}, \mathrm{delivered}\}$ indicates whether the request has not been picked up, has been picked up by a vehicle, or has been delivered.
\end{definition}

In the \problem scenario, the planning horizon is discretized into time slices $t = {0, \ldots, T}$. At the initial time slice $t = 0$, all vehicles are located at their initial origin stations, and their available space equals their full capacity. At each time slice, a scheduler performs the following actions:
\begin{enumerate}[leftmargin=*]
\item For every visible request in the $\mathrm{unassigned}$ state at time $t$ (\ie $\mathrm{time}_m \le t$), assign a vehicle that is currently at the same station and has sufficient space (${spa}_k \ge {vol}_m$) to load the request's cargo. Update the state of the request to $\mathrm{picked}$ and reduce the vehicle's available space accordingly: ${spa}_k := {spa}_k - {vol}_m$.
\item For each vehicle at a station ($dist_k = 0$), select its next destination station and dispatch it. Set the vehicle's travel distance ${dist}_k$ to the distance between the current station and the selected destination.
\item For every vehicle in transit ($dist_k \neq 0$), decrease its remaining travel distance by one unit: ${dist}_k := {dist}_k - 1$.
\item For each vehicle arriving at its destination ($dist_k = 0$), unload all cargos whose destination matches the current station.
Update the state of each delivered request to $\mathrm{delivered}$ and restore the vehicle's available space: ${spa}_k := {spa}_k + {vol}_m$ for each delivered request.
\end{enumerate}
In this scenario, cargo delivery requests emerge dynamically over time, characterizing \problem as a \emph{dynamic} pickup-and-delivery problem.

A \problem instance terminates when the planning horizon reaches the upper limit $T$. The goal of scheduling is to maximize the total profit of requests completed within the time limit while minimizing the aggregate travel cost. 
\subsection{Markov Decision Process (MDP)}

\problem is a sequential decision-making problem over time, and we model its decision process as a Markov decision process, defined as follows:

\begin{definition}[Observation/State]
Since we have a centralized decision system, the system can fully observe all state information at the current time step. The state at time step $t$ includes the distance between stations $e_{n_i, n_j}$, all vehicle information $v_k$, and all appeared requests $r_m$ where $time_m \le t$. 
Notably, within a centralized decision-making system, the relationship \( rel_{v_k, r_m} \in \{\mathrm{unassigned}, \mathrm{picked}, \mathrm{delivered}\} \) between each vehicle \( v_k \) and request \( r_m \) at time \( t \) must be explicitly tracked. This relationship indicates the request has not been picked up, has been picked up by the vehicle, or has been delivered by the vehicle.
\end{definition}

\begin{definition}[Action]
The decision includes decisions for requests and decisions for vehicles.
\begin{itemize}[leftmargin=*]
  \item Request decisions. For the set of $\mathrm{unassigned}$ requests $\mathcal{R}^t = \{r_m \mid state_m=\mathrm{unassigned} \wedge time_m \le t\}$, we need to decide which vehicle to assign them to. The assigned vehicle's current location must be the same as the request's origin, i.e.,
    \begin{equation}
      A^t_{r_m}\in \{v_k \mid dist_k=0 \land to_k=from_m \} \cup \{\tau\} ,
      \label{eq:tau}
    \end{equation}
    where $\tau$ denotes the action of temporarily deferring the request assignment. The assignment must ensure that the vehicle's capacity meets the requirements.
  \item Vehicle decisions. For the set of vehicles that have reached their destinations $\mathcal{V}^t = \{v_k \mid dist_k=0\}$, we need to decide their next destination station, i.e.,
    \begin{equation}
      A^t_{v_k} \in \{n_1, \dots, n_I\} .
    \end{equation}
\end{itemize}
The final joint action space is the Cartesian product of all sub-action spaces:
\begin{equation}
  A^t = \prod \left( 
    \left\{ A^t_{r_m} \mid r_m \in \mathcal{R}^t \right\} 
    \cup 
    \left\{ A^t_{v_k} \mid v_k \in \mathcal{V}^t \right\} 
  \right)
\end{equation}
\end{definition}

\begin{definition}[Transition]
For vehicles $v_k \in \mathcal{V}^t$, we need to update their destination station $to_k$ and remaining time to reach the new destination station $dist_k$ based on $A^t_{v_k}$, and update their remaining space based on unloaded and newly loaded goods. For vehicles $v_k \notin \mathcal{V}^t$, the remaining time to reach their destination stations decreases by one unit. For all requests assigned to vehicles, their $state_m$ and $rel_{v_k, r_m}$ changes to $\mathrm{picked}$, and for all requests that have reached their destinations, their $state_m$ and $rel_{v_k, r_m}$ changes to $\mathrm{delivered}$.
\end{definition}

\begin{definition}[Objective/Reward]
The objective is to optimize the overall routing solution quality by maximizing the profit from completed requests while accounting for vehicle travel costs. Formally, we define the objective at time $T$ as:
\begin{equation}
    obj_{T} = \sum_{t < T} \sum_{k} \left(\sum_{m \in D^t_k} val_m - cost \cdot e_{to^t_k, to^{t+1}_k}\right),
\end{equation}
where $D^t_k$ denotes the set of requests delivered by vehicle $v_k$ by time $t$, and $cost$ represents the cost per unit distance. The single-step reward at each decision point is then defined as the incremental change in the objective value:
\begin{equation}
  rwd_t = obj_t - obj_{t-1}.
  \label{eq:rwd}
\end{equation}
\end{definition}

\section{Methodology}
\subsection{System Overview}
\begin{figure*}[t]
  \centering
  \includegraphics[page=1, width=\linewidth]{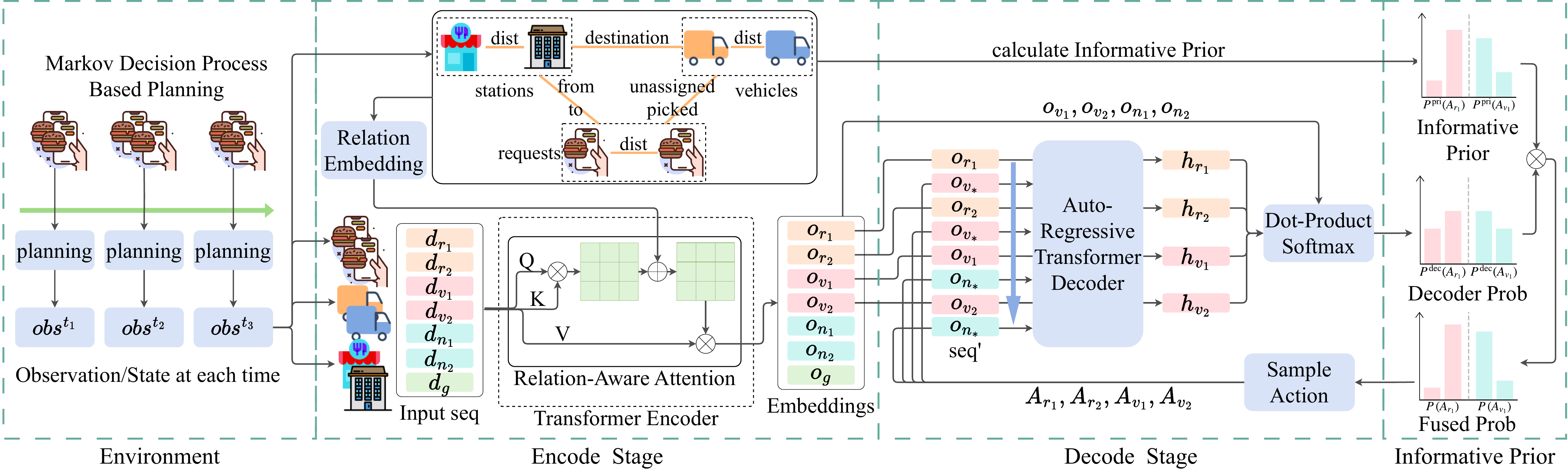}
  \caption{The framework of MAPT. The blue arrow indicates that the elements in the sequence are generated in an AutoRegressive manner. The elements marked with * are the actions that need to be decoded.}
  \label{fig:encoder_decoder}
\end{figure*}

We propose the Multi-Agent Pointer Transformer (MAPT), which is built on the Transformer \cite{NIPS2017_3f5ee243} Encoder-Decoder architecture.
At each time step of the MDP, the current state is fed into MAPT for decision-making.
The Encoder processes the states of all entities and incorporates a Relation-Aware Attention module to capture inter-entity relationships. The Decoder generates actions for each entity in an AutoRegressive manner. To enhance exploration during reinforcement learning, we integrate informative priors into the action selection process. The overall architecture is illustrated in Figure \ref{fig:encoder_decoder}.

\subsection{Encoder}

\subsubsection{Feature Representation with Dense Vectors}

First, we transform the raw input into dense vectors for unified representation. We compute some augmented inputs to enhance the model's perception capabilities, namely \(ori_i\) and \(dest_i\), which represent the number of requests originating from station \(n_i\), and the number of requests destined for station \(n_i\), respectively. Next, we concatenate the raw inputs of stations, requests, and vehicles and pass them through a linear layer to transform them into dense vectors:
\begin{gather}
\bm{d}_{n_i} = \left[ori_i \Vert dest_i\right]\bm{W}^{n}, \quad \bm{d}_{n_i} \in \mathbb{R}^{hs} \\
\bm{d}_{v_k} = \left[cap_k \Vert spa_k \Vert dist_k\right]\bm{W}^{v}, \quad \bm{d}_{v_k} \in \mathbb{R}^{hs} \\
\bm{d}_{r_m} = \left[val_m \Vert vol_m\right]\bm{W}^{r}, \quad \bm{d}_{r_m} \in \mathbb{R}^{hs}
\end{gather}
where \(\Vert\) denotes the concatenation operation for scalars, and $hs$ is the model's hidden size. Additionally, we transform global information into a dense vector, where \(m_t\) represents the current number of requests, as follows:
\begin{equation}
  \bm{d}_g = \left[t \Vert m_t\right]\bm{W}^g, \quad \bm{d}_g \in \mathbb{R}^{hs} .
\end{equation}

\subsubsection{Relation-Aware Attention}

\begin{table}[t]
  \centering
  \tabcolsep=0.10cm
  
  {
    \fontsize{9pt}{10.8pt}\selectfont
  \begin{tabular}{c|c|c|c}
    \toprule
    Relation & Vehicle & Request & Station \\
    \midrule
    Vehicle & distance & \makecell{unassigned \\ picked \\ delivered} & \makecell{is destination \\ isn't destination} \\
    \midrule
    Request & \makecell{unassigned \\ picked \\ delivered} & distance & \makecell{from \\ to} \\
    \midrule
    Station & \makecell{is destination \\ isn't destination} & \makecell{from \\ to} & distance \\
    \bottomrule
  \end{tabular}
  }
  \caption{Relation Types between Vehicle, Request and Station.}
  \label{tab:relation_types}
\end{table}

Although there are some neural network models for solving the Vehicle Routing Problem, these models cannot effectively capture the relationships between entities. To overcome these limitations, this study designs a Relation-Aware Attention module aimed at incorporating relationship information between entities into the embeddings.
For any two entities (stations, requests, vehicles), there are some relationships between them, as shown in Table \ref{tab:relation_types}.
We design a relation encoding network \(\text{Relation-Embedding}(u, v) \in \mathbb{R}\). If the relationship between entities \(u\) and \(v\) is a distance relationship, we use a linear layer to project it; for other relationships, we use a learnable parameter as their embedding.
We denote the relationship matrix between entities as \(\bm{R} \in \mathbb{R}^{|\bm{Q}|\times|\bm{K} |}\), which is calculated as follows:
\begin{equation}
  \bm{R_{u, v}} = \text{Relation-Embedding}(q_u, k_v) ,
\end{equation}
where \(q_u\) represents the \(u\)-th entity in Query sequence ($\bm{Q}$), and \(k_v\) represents the \(v\)-th entity in Key sequence ($\bm{K}$).
Following the definition of the standard Transformer \cite{NIPS2017_3f5ee243}, we represent the Query sequence as \(\bm{Q}\), the Key sequence as \(\bm{K}\), the Value sequence as \(\bm{V}\) and the head dimension as \(d_k\). The Relation-Aware Attention is expressed as:
{
    \fontsize{9pt}{10.8pt}\selectfont
\begin{equation}
  \text{Rel-Aware-Attn}(\bm{Q} ,\bm{K}, \bm{V}, \bm{R}) = \mathrm{softmax}\left(\frac{\bm{QK}^T + \bm{R}}{\sqrt{d_k}}\right)\bm{V} .
\end{equation}
}

We incorporate relationship information into the attention mechanism, enabling the model to capture relationships between entities when generating entity embeddings.

\subsubsection{Encode with Relation-Aware Attention}

We use a standard Transformer Encoder and replace its attention layer with Relation-Aware Attention.
We concatenate the dense vectors of all stations, requests, vehicles, and global information into a sequence and input them into the Encoder to obtain their respective embeddings:
{
    \fontsize{9pt}{10.8pt}\selectfont
\begin{multline}
  [\bm{o}_{r_1}, \dots, \bm{o}_{r_M}, \bm{o}_{v_1}, \dots, \bm{o}_{v_K}, \bm{o}_{n_1}, \dots, \bm{o}_{n_I}, \bm{o}_g] = \\
  \mathrm{Encoder}([\bm{d}_{r_1}, \dots, \bm{d}_{r_M}, \bm{d}_{v_1}, \dots, \bm{d}_{v_K}, \bm{d}_{n_1}, \dots, \bm{d}_{n_I}, \bm{d}_g]) .
\end{multline}
}

\subsection{Decoder}

\subsubsection{Decoder Model}
We use a standard Transformer Decoder and replace its attention layer with Relation-Aware Attention. The cross-attention layer needs to perform cross-attention on the output of the Encoder to perceive all information about the current problem. Additionally, we add learnable positional encoding before the Decoder.

\subsubsection{AutoRegressive Decoding}
We model the joint probability distribution \(A^t\) using chain rule decomposition, which is order-agnostic:
{
    \fontsize{8pt}{9.6}\selectfont
\begin{multline}
  P(A^t) = \prod_{m}^{M}P(A^t_{r_m} \mid A^t_{r_1 \dots r_{m-1}}) \\ \times \prod_{k}^{K}P(A^t_{v_k} \mid A^t_{v_1 \dots v_{k-1}}, A^t_{r_1 \dots r_M}).
  \label{eq:joint_prob}
\end{multline}
}%
From this formula, we can see that each request needs to refer to the decision results of previous requests when making decisions; each vehicle needs to refer to the decision results of previous vehicles and all requests when making decisions.
Due to this sequential decision-making process, we adopt an AutoRegressive approach to generate the decision sequence and model it in a sequence-to-sequence framework, as also observed by \citet{NEURIPS2022_69413f87}.
We combine the Transformer's ability to capture long-distance dependencies with the Pointer Network's \cite{vinyals2015pointer} ability to handle entity selection, using the Transformer Decoder to decode actions.

We detail our decoding process, for notational convenience, we omit the superscript time index \(t\) when unambiguous.
Let our sequence be \(\mathrm{seq}\), initially empty. When decoding the action for the \(m\)-th request \(r_m\), we first append it into the sequence, which then becomes:
\begin{equation}
  \mathrm{seq}=[r_1, A_{r_1},\ \dots, r_m] .
\end{equation}
We first convert entities in the sequence to their corresponding embeddings to obtain $\mathrm{seq}'$ (e.g., converting request $r_m$ to its embedding $\bm{o}_{r_m}$ which is output of the Transformer Encoder).
The \(\mathrm{Decoder}(\mathrm{seq}')\) produces a hidden sequence of the same length as $\mathrm{seq}'$. We take the last item of the hidden sequence, \(\bm{h}_{r_m}\), as the query vector and use the embeddings of all vehicles as the key vectors.
By calculating the attention scores between them, we can quantify the request's preference for each vehicle:
{
\begin{equation}
  P^{\mathrm{dec}}(A_{r_m}=v_k) = \mathrm{softmax}_k(\bm{h}_{r_m}^\top [\bm{o}_{v_1}, \dots, \bm{o}_{v_K}]) ,
\end{equation}
}%
where $\mathrm{softmax}_k(\cdot)$ denotes the $k$-th element of the softmax output.
We determine \(A_{r_m}\) by sampling from \(P^{\mathrm{dec}}(A_{r_m})\) or selecting the maximum probability vehicle. This decoding result is then placed at the end of \(\mathrm{seq}\), providing critical reference information for subsequent decisions:
\begin{equation}
  \mathrm{seq}=[r_1, A_{r_1},\ \dots,\ r_m, A_{r_m}] .
\end{equation}
This process can be repeated to decode all requests. 
We decode all vehicles in the same manner:
{
\begin{equation}
  P^{\mathrm{dec}}(A_{v_k}=n_i) = \mathrm{softmax}_i(\bm{h}_{v_k}^\top [\bm{o}_{n_1}, \dots, \bm{o}_{n_I}]) .
\end{equation}
}%
After all decoding is completed, the elements in the sequence are:
\begin{equation}
  \mathrm{seq} = [r_1, A_{r_1},\ \dots,\ v_1, A_{v_1},\ \dots,\ v_K, A_{v_K}] .
\end{equation}

\subsection{Informative Priors}
\label{sec:heuristic}
The joint action space in \problem is often exponentially large because of multiple vehicles, requests, and stations, which prevents reinforcement learning from exploring useful actions.
To address this, we designed manually-computed informative priors for vehicle selection and destination selection.
By multiplying the probabilities output by the Decoder with informative priors, we guide the agent to prioritize actions with higher potential rewards.

\paragraph{Informative Priors for Vehicle Selection}
In a multi-vehicle scenario, the balance of vehicle load is crucial for overall delivery efficiency and resource utilization. Based on this, we design a load-balancing informative prior, which is inversely proportional to the current vehicle load. The smaller the current load of a vehicle, the higher its informative prior weight in vehicle selection, thereby guiding the assignment of requests toward load balancing:
\begin{equation}
  P^{\mathrm{pri}}(A_{r_m}=v_k) = \frac{spa_k}{cap_k} .
\end{equation}

Although temporarily deferring request assignments may yield positive benefits (e.g., when the current vehicle load is high), we aim to avoid excessive occurrences of such cases, as this would severely affect the exploration efficiency during reinforcement learning. Therefore, we set a small probability for action $\tau$ (defined in Eq.~(\ref{eq:tau})):
\begin{equation}
  P^{\mathrm{pri}}(A_{r_m}=\tau) = \beta .
  \label{eq:beta}
\end{equation}

The vehicle selection probability after informative priors' guidance is:
\begin{equation}
  P(A_{r_m}=v_k) = P^{\mathrm{dec}}(A_{r_m}=v_k) \cdot P^{\mathrm{pri}}(A_{r_m}=v_k) .
\end{equation}

\paragraph{Informative Priors for Destination Selection}
For the destination station selection of vehicle \(v_k\), the informative prior \( P^{\mathrm{pri}}(A_{v_k}=n_i) \) is determined as follows:
\begin{itemize}[leftmargin=*]
    \item If station \(n_i\) has requests to be delivered, \( P^{\mathrm{pri}}(A_{v_k}=n_i) = 1 \).
    \item If station \(n_i\) has requests to be picked up, \( P^{\mathrm{pri}}(A_{v_k}=n_i) = \frac{0.1 \times \overline{\bm{E}}}{e_{to_k, n_i}} \), where \(\overline{\bm{E}}\) denotes the average value of all elements in the distance matrix \(\bm{E}\). This formulation prioritizes stations that are closer in distance. Since the number of stations with pickup requests is significantly larger than those with delivery requests, we use a smaller coefficient (0.1) to balance the vehicle's priorities between pickup and delivery tasks.  
    \item For stations \(n_i\) that do not fall into the above two categories, \( P^{\mathrm{pri}}(A_{v_k}=n_i) = 0 \).
\end{itemize}
The destination selection probability after informative priors' guidance is:
\begin{equation}
  P(A_{v_k}=n_i) = P^{\mathrm{dec}}(A_{v_k}=n_i) \cdot P^{\mathrm{pri}}(A_{v_k}=n_i) .
\end{equation}

\subsection{Optimization via PPO}

We use Proximal Policy Optimization \cite{schulman2017proximal} to train MAPT.
Our value function is defined as $V(s_t) = \text{MLP}(\bm{o_g})$, 
where $\text{MLP}$ transforms global information into a scalar. We use Generalized Advantage Estimation \cite{schulman2015high} to balance the bias and variance of advantage estimation:
\begin{equation}
  \hat{A}_t^{\mathrm{GAE}} = \sum_{l = 0}^{\infty}(\gamma\lambda)^l(rwd_{t + l}+\gamma V(s_{t + l+1})-V(s_{t + l})).
  \label{eq:gae}
\end{equation}
The Actor loss is defined as
{
  \fontsize{9pt}{10.8}\selectfont
\begin{gather}
  \rho_t(\theta)=\frac{\pi_\theta(A^t|s_t)}{\pi_{\theta_{\mathrm{old}}}(A^t|s_t)}, \\
  \begin{split}
    \mathcal{L}^{\mathrm{CLIP}}(\theta) &= \mathbb{E}_{t} \Big[ \min \big( 
    \rho_t(\theta) \hat{A}_t^{\mathrm{GAE}}, \\
    &\qquad \qquad \mathrm{CLIP}(\rho_t(\theta), 1-\epsilon, 1+\epsilon) \hat{A}_t^{\mathrm{GAE}} \big) \Big],
\end{split}
  \label{eq:actor_loss}
\end{gather}
}%
where $\pi_\theta(A^t|s_t)$ is the joint probability defined in Eq.~(\ref{eq:joint_prob}). The Critic loss is defined as
\begin{equation}
  \mathcal{L}^{\mathrm{Critic}}=\left(V(s_t)-(r_{t + 1}+\gamma V(s_{t + 1}))\right)^2.
  \label{eq:critic_loss}
\end{equation}
The final total loss is
\begin{equation}
  \mathcal{L} = \mathcal{L}^{\mathrm{CLIP}}+\alpha \mathcal{L}^{\mathrm{Critic}}.
  \label{eq:total_loss}
\end{equation}

\section{Experiments}

\begin{table*}[h]
  \tabcolsep=0.06cm
  \centering
{
  \fontsize{9pt}{10.8pt}\selectfont
  \begin{tabular}{c|cc|cc|cc|cc|cc|cc|cc|cc}
    \toprule
    Scenario & \multicolumn{2}{c|}{synth-S} & \multicolumn{2}{c|}{synth-S-cost} & \multicolumn{2}{c|}{synth-L} & \multicolumn{2}{c|}{synth-L-cost} & \multicolumn{2}{c|}{dhrd-tpe} & \multicolumn{2}{c|}{dhrd-sg} & \multicolumn{2}{c|}{dhrd-se} & \multicolumn{2}{c}{synth-XL} \\
    Metric & Obj.$\uparrow$ & Comp.$\uparrow$ & Obj.$\uparrow$ & Comp.$\uparrow$ & Obj.$\uparrow$ & Comp.$\uparrow$ & Obj.$\uparrow$ & Comp.$\uparrow$ & Obj.$\uparrow$ & Comp.$\uparrow$ & Obj.$\uparrow$ & Comp.$\uparrow$ & Obj.$\uparrow$ & Comp.$\uparrow$ & Obj.$\uparrow$ & Comp.$\uparrow$ \\
    \midrule

OR Tools & 182.1 & 0.54 & 117.4 & 0.52 & N/A & N/A & N/A & N/A & N/A & N/A & N/A & N/A & 55.3 & 0.45 & N/A & N/A \\
SA & 162.5 & 0.52 & 108.1 & 0.52 & 641.1 & 0.28 & 339.3 & 0.29 & 244.9 & 0.31 & 136.7 & 0.30 & 50.4 & 0.41 & 1531.9 & 0.31 \\
GA & 108.7 & 0.30 & 58.3 & 0.29 & 329.4 & 0.14 & 89.5 & 0.12 & 64.1 & 0.08 & 40.9 & 0.13 & 21.3 & 0.17 & 863.8 & 0.17 \\
\midrule
SA\textsuperscript{*} & 133.0 & 0.42 & 81.2 & 0.42 & 563.1 & 0.25 & 284.8 & 0.24 & 223.3 & 0.28 & 139.6 & 0.30 & 56.6 & 0.46 & 1650.9 & 0.34 \\
GA\textsuperscript{*} & 75.2 & 0.21 & 39.4 & 0.19 & 219.9 & 0.09 & 88.0 & 0.09 & 53.8 & 0.07 & 39.0 & 0.13 & 18.4 & 0.15 & 543.2 & 0.11 \\
MAPDP\textsuperscript{*} & 174.9 & 0.53 & 82.0 & 0.49 & 177.6 & 0.08 & 65.2 & 0.08 & 287.5 & 0.36 & 247.2 & 0.35 & 72.9 & 0.56 & 807.0 & 0.16 \\
PARCO\textsuperscript{*} & 178.8 & 0.53 & 86.6 & 0.52 & 177.6 & 0.08 & 65.2 & 0.08 & 391.2 & 0.49 & 267.5 & 0.46 & 64.2 & 0.49 & 1834.7 & 0.36 \\
\midrule
Nearest & 189.2 & 0.57 & 124.1 & 0.57 & 962.6 & 0.41 & 592.3 & 0.41 & 309.2 & 0.39 & 194.5 & 0.40 & 39.8 & 0.32 & 2641.1 & 0.53 \\
\midrule
MAPDP & 157.5 & 0.41 & 70.0 & 0.42 & 958.1 & 0.37 & 385.2 & 0.37 & 135.4 & 0.17 & 64.9 & 0.18 & 19.3 & 0.16 & 2543.1 & 0.50 \\
MAPT & \textbf{275.2} & \textbf{0.83} & \textbf{192.1} & \textbf{0.84} & \textbf{1875.1} & \textbf{0.80} & \textbf{1348.6} & \textbf{0.81} & \textbf{697.2} & \textbf{0.87} & \textbf{376.1} & \textbf{0.71} & \textbf{78.9} & \textbf{0.64} & \textbf{3227.5} & \textbf{0.65} \\

\bottomrule
\end{tabular}
}
  \caption{Overall performance comparison on synthetic datasets and real-world datasets. The best result in each column is bolded. "N/A" indicates that the algorithm cannot provide results within an acceptable time frame.}
  \label{tab:compare_all}
\end{table*}

\begin{table*}[t]
  \centering
  \tabcolsep=0.06cm
{
  \fontsize{9pt}{10.8pt}\selectfont
  \begin{tabular}{c|cc|cc|cc|cc|cc|cc|cc|cc}
    \toprule
    Scenario & \multicolumn{2}{c|}{synth-S} & \multicolumn{2}{c|}{synth-S-cost} & \multicolumn{2}{c|}{synth-L} & \multicolumn{2}{c|}{synth-L-cost} & \multicolumn{2}{c|}{dhrd-tpe} & \multicolumn{2}{c|}{dhrd-sg} & \multicolumn{2}{c|}{dhrd-se} & \multicolumn{2}{c}{synth-XL} \\
    Metric & Obj.$\uparrow$ & Comp.$\uparrow$ & Obj.$\uparrow$ & Comp.$\uparrow$ & Obj.$\uparrow$ & Comp.$\uparrow$ & Obj.$\uparrow$ & Comp.$\uparrow$ & Obj.$\uparrow$ & Comp.$\uparrow$ & Obj.$\uparrow$ & Comp.$\uparrow$ & Obj.$\uparrow$ & Comp.$\uparrow$ & Obj.$\uparrow$ & Comp.$\uparrow$ \\
    \midrule
MAPT & \textbf{275.2} & \textbf{0.83} & \textbf{192.1} & \textbf{0.84} & \textbf{1875.1} & \textbf{0.80} & \textbf{1348.6} & \textbf{0.81} & \textbf{697.2} & \textbf{0.87} & \textbf{376.1} & \textbf{0.71} & \textbf{78.9} & \textbf{0.64} & \textbf{3227.5} & \textbf{0.65} \\
w/o Rel & 270.9 & 0.82 & 190.7 & 0.83 & 1869.9 & 0.80 & 1338.4 & 0.81 & 692.1 & 0.87 & 365.3 & 0.69 & 77.1 & 0.63 & 3136.8 & 0.63 \\
w/o AR & 217.2 & 0.65 & 171.1 & 0.77 & 1604.0 & 0.68 & 1074.5 & 0.69 & 611.4 & 0.76 & 258.1 & 0.50 & 49.4 & 0.40 & 1905.8 & 0.38 \\
w/o Priors & 175.2 & 0.54 & 92.7 & 0.54 & 1089.5 & 0.47 & 529.1 & 0.48 & 511.6 & 0.64 & 250.9 & 0.49 & 56.6 & 0.46 & 1375.0 & 0.27 \\
\bottomrule
\end{tabular}
}
  \caption{Results of ablation studies. The best result in each column is bolded.}
  \label{tab:ablation}
\end{table*}

\subsection{Experimental Scenarios}
To validate the effectiveness of the MAPT framework, we conducted experiments on 8 datasets, including both synthetic and real-world datasets, for comprehensive evaluation.
A brief description of the datasets is provided below.
\begin{itemize}[leftmargin=*]
  \item {\em Synthetic Dataset}: We generated synthetic datasets with different scales of stations $I \in \{20, 50, 300\}$, corresponding to dataset suffixes \texttt{-S}, \texttt{-L}, and \texttt{-XL}, respectively. For these datasets, the number of requests and vehicles were set as $(M, K) = (110, 5)$, $(550, 15)$, and $(550, 50)$, respectively. Vehicle capacity was fixed at 3. Travel cost per unit distance is set to 0 or 0.3, with the latter case marked by an additional suffix \texttt{-cost}. Requests were uniformly sampled from stations, with appearance times sampled from $\mathrm{Uniform}(1, T)$, where $T$ was set to 58 or 128. The profit for each request was the distance between its origin and destination.
  \item {\em Real-World Dataset}: We used the DHRD dataset \cite{assylbekov2023delivery}, which contains food delivery requests from three cities: Taipei (\texttt{tpe}), Singapore (\texttt{sg}), and Stockholm (\texttt{se}). Each city was treated as an independent dataset with $I = 36$ stations. The corresponding numbers of requests and vehicles were $(M, K) = (800, 20)$, $(700, 15)$, and $(200, 3)$ for the three cities, respectively. Vehicle capacity was set to 6. The dataset was divided into 76 days for training/validation and 14 days for testing.
\end{itemize}

\subsection{Performance Evaluation}

\subsubsection{Baselines}
We evaluate our MAPT against several baselines, including Rolling-Horizon algorithms, static algorithms, rule-based algorithms, and MDP-based algorithms.
\begin{itemize}[leftmargin=*]
\item \textbf{Rolling-Horizon Algorithms:}
The Rolling-Horizon paradigm addresses dynamic problems by decomposing them into static subproblems within a sliding time window. For solving these static subproblems, we employ \textbf{OR Tools}, \textbf{Simulated Annealing (SA)}, and \textbf{Genetic Algorithm (GA)}.
\item \textbf{Static Algorithms:}
These methods assume complete foreknowledge of all requests (though unrealistic), solving the problem statically without Rolling-Horizon. \textbf{MAPDP*} \cite{zong2022mapdp} is a MARL approach with Encoder-Decoder for static problems, adapted to handle distance matrix inputs. \textbf{PARCO*} \cite{berto2024parco} is a Transformer-based RL framework with parallel decision-making, modified to support distance matrix inputs. \textbf{SA*/GA*} are static versions of our Rolling-Horizon SA and GA with full request information.
\item \textbf{Rule-Based Algorithms:}
These methods use predefined rules for scheduling decisions. \textbf{Nearest} is a greedy approach that always selects the closest available request for pickup or delivery.
\item \textbf{MDP-Based Algorithms:}
At each step of the MDP, we solve the problem statically using \textbf{MAPDP} \cite{zong2022mapdp} as described in the Static Algorithms section, and only execute the first step of the solution to ensure applicability to dynamic problems.
\end{itemize}

\subsubsection{Overall Comparison}

The performance results on synthetic and real-world datasets are summarized in Table \ref{tab:compare_all}, using the objective (Obj.) and request completion rate (Comp.) as metrics (higher is better).
MAPT significantly outperforms all baselines across all datasets. MAPT also surpasses the state-of-the-art MAPDP, which fails to model multi-vehicle joint probabilities and lacks our informative priors.
As can be seen from Table \ref{tab:compare_time}, the decision time of MAPT is significantly shorter than that of exact and meta-heuristic algorithms.

\begin{table}[t]
  \centering
  \tabcolsep=0.04cm
{
  \fontsize{9pt}{10.8pt}\selectfont
  \begin{tabular}{cccccc}
    \toprule
    Scenario & synth-S & synth-S-cost & synth-L & synth-L-cost & synth-XL \\
    \midrule
OR Tools & 401.727 & 459.143 & N/A & N/A & N/A \\
SA & 23.684 & 23.255 & 166.660 & 164.102 & 613.920 \\
GA & 16.654 & 16.448 & 103.147 & 99.864 & 354.789 \\
Nearest & 0.059 & 0.060 & 0.117 & 0.119 & 0.148 \\
MAPDP & 0.794 & 0.827 & 31.981 & 32.544 & 33.217 \\
MAPT & 1.420 & 1.416 & 8.300 & 7.888 & 27.434 \\
    \bottomrule
\end{tabular}
}
\caption{Comparison of inference times on the synthetic dataset (in seconds), which is the average running time of each instance in the dataset.
  }
  \label{tab:compare_time}
\end{table}

\subsection{Ablation Studies}

We conducted ablation studies on all datasets to validate the effectiveness of key components:
\begin{itemize}[leftmargin=*]
  \item \textbf{w/o Relation-Aware Attention (Rel)}: Replaces the Relation-Aware Attention module with the original attention mechanism.
  \item \textbf{w/o AutoRegressive Decoding (AR)}: Decodes each action independently and employs a conflict handler to resolve conflicts where request assignments exceed vehicle capacity.
  \item \textbf{w/o Informative Priors (Priors)}: Removes informative priors guidance for vehicle and destination selection.
\end{itemize}

The results are shown in Table \ref{tab:ablation}. As can be seen from the table, AutoRegressive Decoding and Informative Priors significantly improve performance across all datasets, and Relation-Aware Attention also contributes to some extent.
The ablation experiments demonstrate that all three proposed components enhance the final performance, validating their effectiveness.

\subsection{Sensitivity Analysis}

\begin{figure}[htbp]
    \centering

    \begin{subfigure}[b]{0.20\textwidth}
        \includegraphics[width=\linewidth]{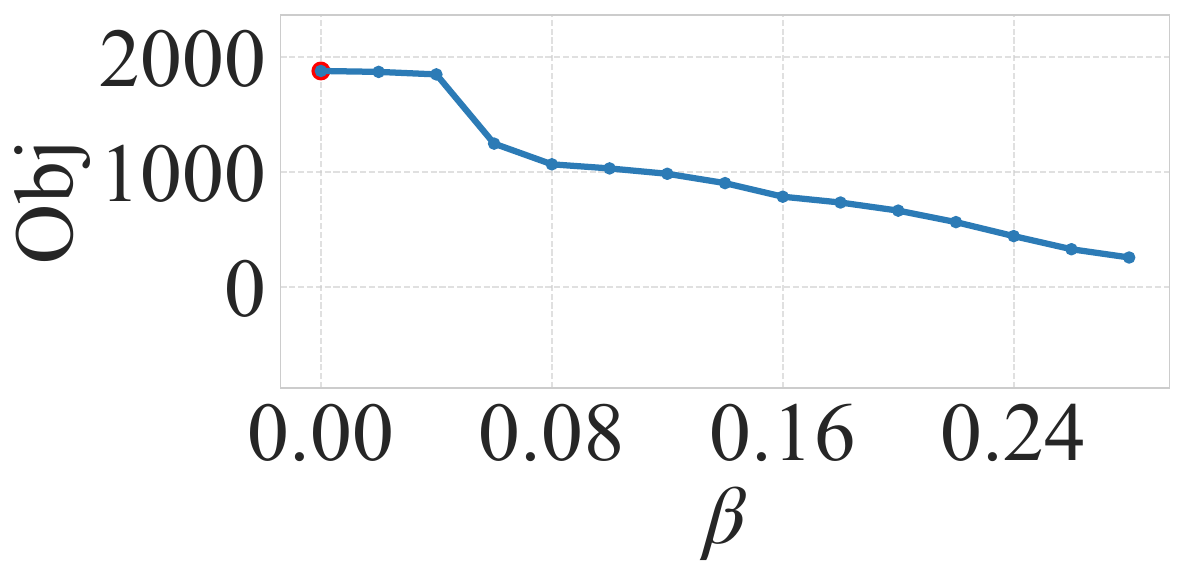}
        \caption{synthetic-L}
    \end{subfigure}
    \begin{subfigure}[b]{0.20\textwidth}
        \includegraphics[width=\linewidth]{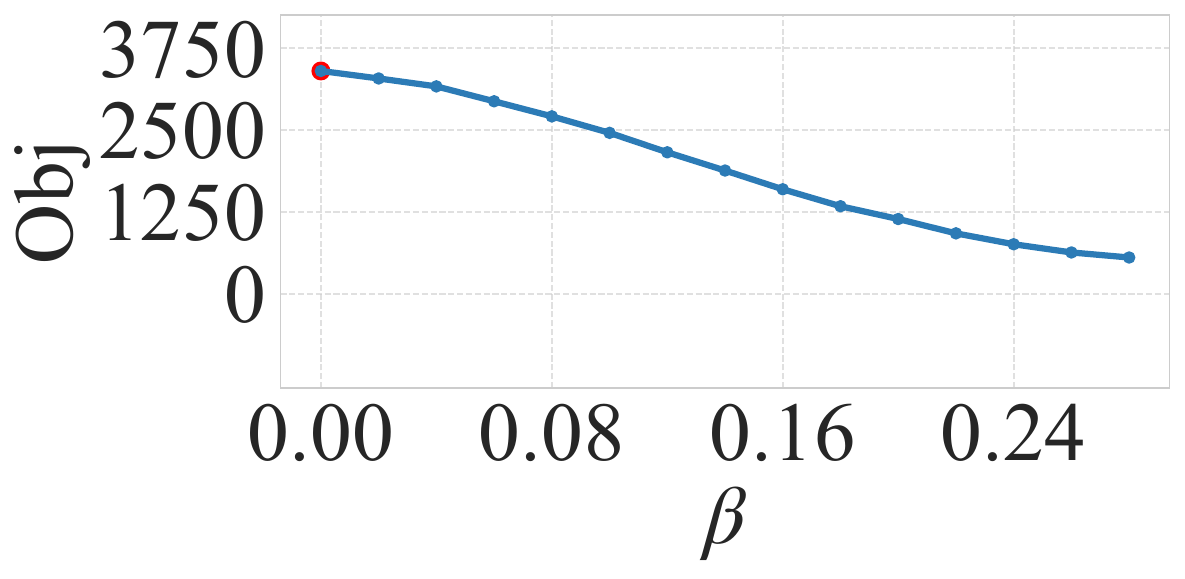}
        \caption{synthetic-XL}
    \end{subfigure}
    \begin{subfigure}[b]{0.20\textwidth}
        \includegraphics[width=\linewidth]{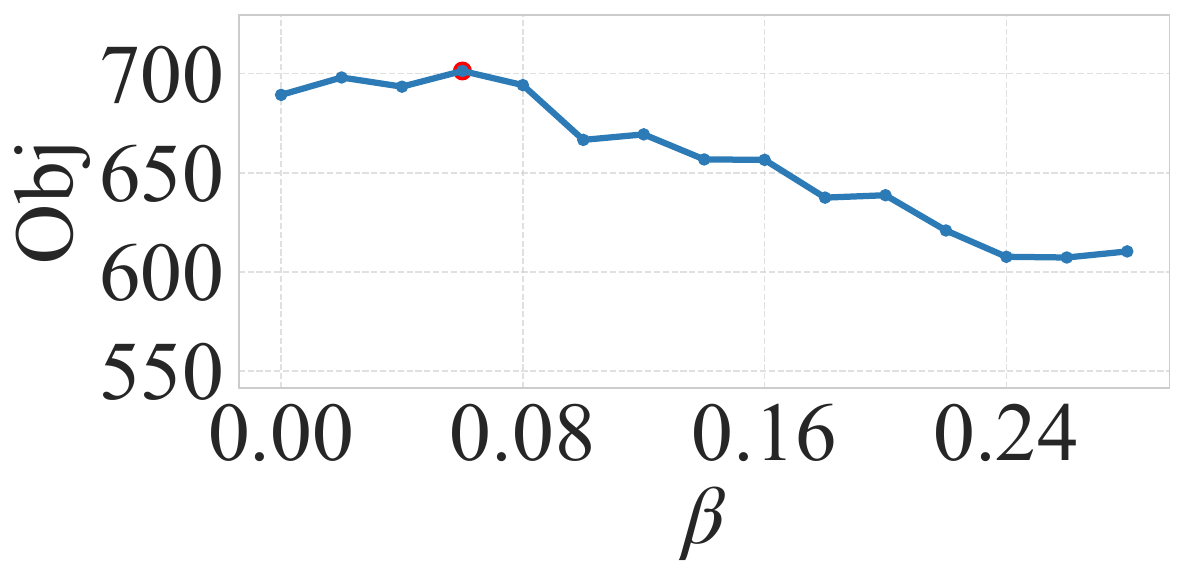}
        \caption{dhrd-tpe}
    \end{subfigure}
    \begin{subfigure}[b]{0.20\textwidth}
        \includegraphics[width=\linewidth]{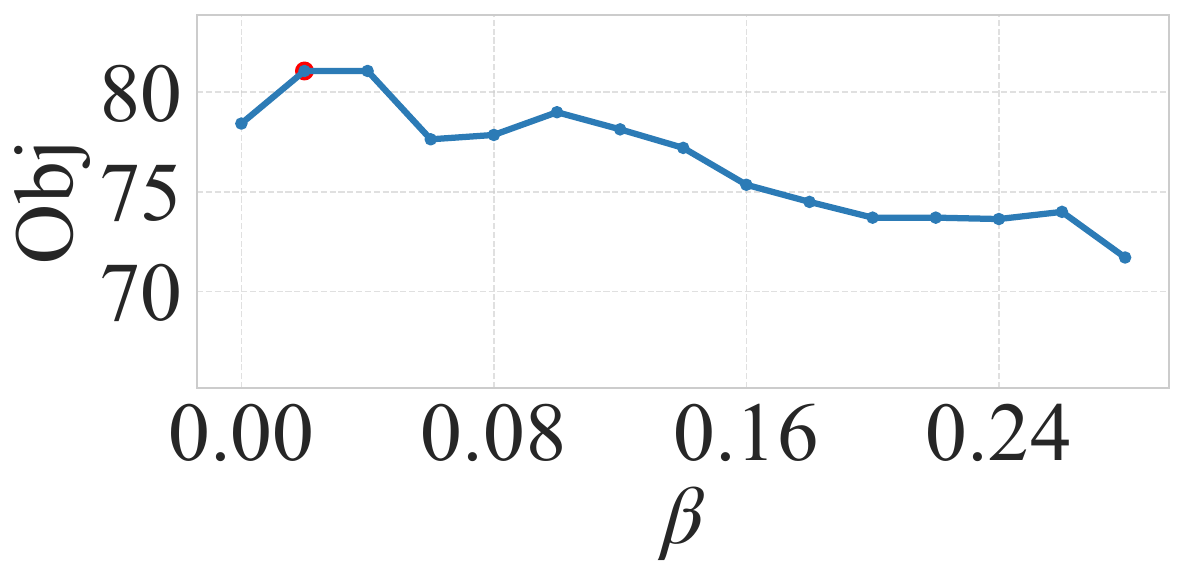}
        \caption{dhrd-se}
    \end{subfigure}
    
    \caption{Sensitivity analysis of $\beta$ across datasets.}
    \label{fig:sense_beta}
\end{figure}

We conduct a sensitivity analysis on the hyperparameter $\beta$ as defined in Eq.~(\ref{eq:beta}).
We vary $\beta$ from 0 to 0.3 in increments of 0.02 and observe the resulting changes in the final objective.
The results on the 4 representative datasets are illustrated in Figure \ref{fig:sense_beta}.
As shown in the figure, when $\beta$ is small, the performance is better and more stable, while larger values of $\beta$ lead to a gradual decline in performance.
However, in the dhrd-tpe and dhrd-se datasets, using a small but non-zero $\beta$ improves performance. This indicates that slightly deferring request allocation can be beneficial in these cases.

\section{Related Work}

\subsection{Dynamic Pickup and Delivery Problems}
Dynamic Pickup and Delivery Problems have been studied through exact, heuristic, metaheuristic, and learning-based methods \cite{cai2023survey}. 
Exact methods such as MILP can yield optimal results but are restricted to small instances \cite{liu2018dynamic, savelsbergh1998drive}. 
Heuristic methods \cite{fikar2018decision, fikar2018decision2, andersson2021comparative} are efficient for large-scale dynamic problems but cannot ensure optimality. 
Metaheuristics \cite{schilde2011metaheuristics, tirado2017improved, novaes2015dynamic} improve scalability but still struggle in highly dynamic settings. 
Reinforcement learning methods have been explored \cite{li2021learning, yu2025bigcity}, but they mainly focus on single-vehicle operations.
Hybrid approaches first learn edge costs using neural networks and then apply traditional optimization techniques \cite{jiang2023continuous, wang2021personalized, wang2019empowering, wang2025gtg, wu2019learning}.
Other studies have addressed ride-on-demand problems with dynamic pricing \cite{guo2019rod, guo2020force, guo2023seeking}, as well as dynamic metro scheduling \cite{wang2022shortening}.
There are also entity representation learning methods that use GNNs, hyper-GNNs, contrastive learning, transfer learning, or LLM enhancement and can be applied to spatio-temporal system optimization \cite{wu2020learning, yang2025hygmap, han2025bridging, jiang2023self, licross, cheng2025poi, zhang2024veccity}.

\subsection{RL-based Methods for Vehicle Routing Problems}
Reinforcement Learning has been applied to both single and multiple Vehicle Routing Problems. 
For single-vehicle problems, attention-based models \cite{vinyals2015pointer, bello2016neural, NEURIPS2018_9fb4651c, kool2018attention} and GNN-based variants \cite{fellek2023deep, heydaribeni2023hypop} have been proposed. 
For multi-vehicle problems, multi-agent RL methods \cite{zhang2020multi, zong2022mapdp, zhang2023coordinated} have been developed but mainly for static settings.
Recent RL studies for dynamic settings \cite{anuar2022multi, xiang2024centralized} have emerged, but they simplify the original problem by introducing additional assumptions.

\section{Conclusion}

This paper presents MAPT
to solve the cooperative Multi-Vehicle Dynamic Pickup and Delivery Problem with Stochastic Requests.
MAPT utilizes the Transformer’s Encoder-Decoder architecture combining with Pointer Network to generate joint action sequences in an AutoRegressive manner.
By incorporating the Relation-Aware Attention module and informative priors, the framework achieves a better performance.
Experiments show that MAPT outperforms existing baselines in solution quality across synthetic datasets and real-world datasets.

\section*{Acknowledgments}
This work is supported by the National Key R\&D Program of China (2023YFC3304700). Prof. Jingyuan Wang's work was partially supported by the National Natural Science Foundation of China (No. 72222022, 72171013) and the Fundamental Research Funds for the Central Universities (JKF-2025017226182). Prof. Junjie Wu’s work was partially supported by the National Natural Science Foundation of China (72242101, 72031001), the Outstanding Young Scientist Program of Beijing Universities (JWZQ20240201002), and the Artificial Intelligence Technology R\&D Center for Exploration and Development of China National Petroleum Corporation (2024-KFKT-22).

\bibliography{aaai2026}

@article{vinyals2015pointer,
  title={Pointer networks},
  author={Vinyals, Oriol and Fortunato, Meire and Jaitly, Navdeep},
  journal={Advances in neural information processing systems},
  volume={28},
  year={2015}
}

@article{bello2016neural,
  title={Neural combinatorial optimization with reinforcement learning},
  author={Bello, Irwan and Pham, Hieu and Le, Quoc V and Norouzi, Mohammad and Bengio, Samy},
  journal={arXiv preprint arXiv:1611.09940},
  year={2016}
}

@inproceedings{NEURIPS2018_9fb4651c,
 author = {Nazari, MohammadReza and Oroojlooy, Afshin and Snyder, Lawrence and Takac, Martin},
 booktitle = {Advances in Neural Information Processing Systems},
 editor = {S. Bengio and H. Wallach and H. Larochelle and K. Grauman and N. Cesa-Bianchi and R. Garnett},
 pages = {},
 publisher = {Curran Associates, Inc.},
 title = {Reinforcement Learning for Solving the Vehicle Routing Problem},
 url = {https://proceedings.neurips.cc/paper_files/paper/2018/file/9fb4651c05b2ed70fba5afe0b039a550-Paper.pdf},
 volume = {31},
 year = {2018}
}

@article{kool2018attention,
  title={Attention, learn to solve routing problems!},
  author={Kool, Wouter and Van Hoof, Herke and Welling, Max},
  journal={arXiv preprint arXiv:1803.08475},
  year={2018}
}

@inproceedings{fellek2023deep,
  title={Deep Graph Representation Learning to Solve Vehicle Routing Problem},
  author={Fellek, Getu and Farid, Ahmed and Gebreyesus, Goytom and Fujimura, Shigeru and Yoshie, Osamu},
  booktitle={2023 International Conference on Machine Learning and Cybernetics (ICMLC)},
  pages={172--180},
  year={2023},
  organization={IEEE}
}

@article{heydaribeni2023hypop,
  title={HypOp: Distributed Constrained Combinatorial Optimization leveraging Hypergraph Neural Networks},
  author={Heydaribeni, Nasimeh and Zhan, Xinrui and Zhang, Ruisi and Eliassi-Rad, Tina and Koushanfar, Farinaz},
  journal={arXiv preprint arXiv:2311.09375},
  year={2023}
}

@article{xiang2024centralized,
  title={Centralized Deep Reinforcement Learning Method for Dynamic Multi-Vehicle Pickup and Delivery Problem With Crowdshippers},
  author={Xiang, Chuankai and Wu, Zhibin and Tu, Jiancheng and Huang, Jun},
  journal={IEEE Transactions on Intelligent Transportation Systems},
  year={2024},
  publisher={IEEE}
}

@article{anuar2022multi,
  title={A multi-depot dynamic vehicle routing problem with stochastic road capacity: An MDP model and dynamic policy for post-decision state rollout algorithm in reinforcement learning},
  author={Anuar, Wadi Khalid and Lee, Lai Soon and Seow, Hsin-Vonn and Pickl, Stefan},
  journal={Mathematics},
  volume={10},
  number={15},
  pages={2699},
  year={2022},
  publisher={MDPI}
}

@article{zhang2020multi,
  title={Multi-vehicle routing problems with soft time windows: A multi-agent reinforcement learning approach},
  author={Zhang, Ke and He, Fang and Zhang, Zhengchao and Lin, Xi and Li, Meng},
  journal={Transportation Research Part C: Emerging Technologies},
  volume={121},
  pages={102861},
  year={2020},
  publisher={Elsevier}
}

@inproceedings{NEURIPS2022_69413f87,
 author = {Wen, Muning and Kuba, Jakub and Lin, Runji and Zhang, Weinan and Wen, Ying and Wang, Jun and Yang, Yaodong},
 booktitle = {Advances in Neural Information Processing Systems},
 editor = {S. Koyejo and S. Mohamed and A. Agarwal and D. Belgrave and K. Cho and A. Oh},
 pages = {16509--16521},
 publisher = {Curran Associates, Inc.},
 title = {Multi-Agent Reinforcement Learning is a Sequence Modeling Problem},
 url = {https://proceedings.neurips.cc/paper_files/paper/2022/file/69413f87e5a34897cd010ca698097d0a-Paper-Conference.pdf},
 volume = {35},
 year = {2022}
}

@inproceedings{zong2022mapdp,
  title={Mapdp: Cooperative multi-agent reinforcement learning to solve pickup and delivery problems},
  author={Zong, Zefang and Zheng, Meng and Li, Yong and Jin, Depeng},
  booktitle={Proceedings of the AAAI conference on artificial intelligence},
  volume={36},
  number={9},
  pages={9980--9988},
  year={2022}
}

@article{berto2024parco,
  title={PARCO: Learning Parallel Autoregressive Policies for Efficient Multi-Agent Combinatorial Optimization},
  author={Berto, Federico and Hua, Chuanbo and Luttmann, Laurin and Son, Jiwoo and Park, Junyoung and Ahn, Kyuree and Kwon, Changhyun and Xie, Lin and Park, Jinkyoo},
  journal={arXiv preprint arXiv:2409.03811},
  year={2024}
}

@article{schulman2017proximal,
  title={Proximal policy optimization algorithms},
  author={Schulman, John and Wolski, Filip and Dhariwal, Prafulla and Radford, Alec and Klimov, Oleg},
  journal={arXiv preprint arXiv:1707.06347},
  year={2017}
}

@article{zhang2023coordinated,
  title={Coordinated multi-agent hierarchical deep reinforcement learning to solve multi-trip vehicle routing problems with soft time windows},
  author={Zhang, Zixian and Qi, Geqi and Guan, Wei},
  journal={IET Intelligent Transport Systems},
  volume={17},
  number={10},
  pages={2034--2051},
  year={2023},
  publisher={Wiley Online Library}
}

@inproceedings{NIPS2017_3f5ee243,
 author = {Vaswani, Ashish and Shazeer, Noam and Parmar, Niki and Uszkoreit, Jakob and Jones, Llion and Gomez, Aidan N and Kaiser, \L ukasz and Polosukhin, Illia},
 booktitle = {Advances in Neural Information Processing Systems},
 editor = {I. Guyon and U. Von Luxburg and S. Bengio and H. Wallach and R. Fergus and S. Vishwanathan and R. Garnett},
 pages = {},
 publisher = {Curran Associates, Inc.},
 title = {Attention is All you Need},
 url = {https://proceedings.neurips.cc/paper_files/paper/2017/file/3f5ee243547dee91fbd053c1c4a845aa-Paper.pdf},
 volume = {30},
 year = {2017}
}

@inproceedings{assylbekov2023delivery,
  title={Delivery Hero Recommendation Dataset: A Novel Dataset for Benchmarking Recommendation Algorithms},
  author={Assylbekov, Yernat and Bali, Raghav and Bovard, Luke and Klaue, Christian},
  booktitle={Proceedings of the 17th ACM Conference on Recommender Systems},
  pages={1042--1044},
  year={2023}
}

@article{lu2004exact,
  title={An exact algorithm for the multiple vehicle pickup and delivery problem},
  author={Lu, Quan and Dessouky, Maged},
  journal={Transportation Science},
  volume={38},
  number={4},
  pages={503--514},
  year={2004},
  publisher={INFORMS}
}

@article{cai2023survey,
  title={A survey of dynamic pickup and delivery problems},
  author={Cai, Junchuang and Zhu, Qingling and Lin, Qiuzhen and Ma, Lijia and Li, Jianqiang and Ming, Zhong},
  journal={Neurocomputing},
  pages={126631},
  year={2023},
  publisher={Elsevier}
}

@inproceedings{liu2018dynamic,
  title={Dynamic scheduling for pickup and delivery with time windows},
  author={Liu, Shudong and Tan, Peng Hui and Kurniawan, Ernest and Zhang, Peng and Sun, Sumei},
  booktitle={2018 IEEE 4th World Forum on Internet of Things (WF-IoT)},
  pages={767--770},
  year={2018},
  organization={IEEE}
}

@article{savelsbergh1998drive,
  title={Drive: Dynamic routing of independent vehicles},
  author={Savelsbergh, Martin and Sol, Marc},
  journal={Operations Research},
  volume={46},
  number={4},
  pages={474--490},
  year={1998},
  publisher={INFORMS}
}

@article{fikar2018decision,
  title={A decision support system to investigate food losses in e-grocery deliveries},
  author={Fikar, Christian},
  journal={Computers \& Industrial Engineering},
  volume={117},
  pages={282--290},
  year={2018},
  publisher={Elsevier}
}

@article{fikar2018decision2,
  title={A decision support system to investigate dynamic last-mile distribution facilitating cargo-bikes},
  author={Fikar, Christian and Hirsch, Patrick and Gronalt, Manfred},
  journal={International Journal of Logistics Research and Applications},
  volume={21},
  number={3},
  pages={300--317},
  year={2018},
  publisher={Taylor \& Francis}
}

@misc{andersson2021comparative,
  title={A Comparative Study on a Dynamic Pickup and Delivery Problem: Improving routing and order assignment in same-day courier operations},
  author={Andersson, Tomas},
  year={2021}
}

@article{sheridan2013dynamic,
  title={The dynamic nearest neighbor policy for the multi-vehicle pick-up and delivery problem},
  author={Sheridan, Patricia Kristine and Gluck, Erich and Guan, Qi and Pickles, Thomas and Balc{\i}og, Bar{\i}{\c{s}} and Benhabib, Beno and others},
  journal={Transportation Research Part A: Policy and Practice},
  volume={49},
  pages={178--194},
  year={2013},
  publisher={Elsevier}
}

@article{schilde2011metaheuristics,
  title={Metaheuristics for the dynamic stochastic dial-a-ride problem with expected return transports},
  author={Schilde, Michael and Doerner, Karl F and Hartl, Richard F},
  journal={Computers \& operations research},
  volume={38},
  number={12},
  pages={1719--1730},
  year={2011},
  publisher={Elsevier}
}

@article{tirado2017improved,
  title={Improved solutions to dynamic and stochastic maritime pick-up and delivery problems using local search},
  author={Tirado, Gregorio and Hvattum, Lars Magnus},
  journal={Annals of Operations Research},
  volume={253},
  pages={825--843},
  year={2017},
  publisher={Springer}
}

@article{novaes2015dynamic,
  title={Dynamic milk-run OEM operations in over-congested traffic conditions},
  author={Novaes, Antonio GN and Bez, Edson T and Burin, Paulo J and Arag{\~a}o Jr, Dmontier P},
  journal={Computers \& Industrial Engineering},
  volume={88},
  pages={326--340},
  year={2015},
  publisher={Elsevier}
}

@inproceedings{geiser2020best,
  title={Best-match in a set of single-vehicle dynamic pickup and delivery problem using ant colony optimization},
  author={Geiser, Thomas and Hanne, Thomas and Dornberger, Rolf},
  booktitle={Proceedings of the 2020 the 3rd International Conference on Computers in Management and Business},
  pages={126--131},
  year={2020}
}

@inproceedings{cai2022efficient,
  title={An efficient multi-objective evolutionary algorithm for a practical dynamic pickup and delivery problem},
  author={Cai, Junchuang and Zhu, Qingling and Lin, Qiuzhen and Li, Jianqiang and Chen, Jianyong and Ming, Zhong},
  booktitle={International conference on intelligent computing},
  pages={27--40},
  year={2022},
  organization={Springer}
}

@article{schulman2015high,
  title={High-dimensional continuous control using generalized advantage estimation},
  author={Schulman, John and Moritz, Philipp and Levine, Sergey and Jordan, Michael and Abbeel, Pieter},
  journal={arXiv preprint arXiv:1506.02438},
  year={2015}
}

@inproceedings{li2021learning,
  title={Learning to optimize industry-scale dynamic pickup and delivery problems},
  author={Li, Xijun and Luo, Weilin and Yuan, Mingxuan and Wang, Jun and Lu, Jiawen and Wang, Jie and L{\"u}, Jinhu and Zeng, Jia},
  booktitle={2021 IEEE 37th international conference on data engineering (ICDE)},
  pages={2511--2522},
  year={2021},
  organization={IEEE}
}

@inproceedings{yu2025bigcity,
  title={BIGCity: A universal spatiotemporal model for unified trajectory and traffic state data analysis},
  author={Yu, Xie and Wang, Jingyuan and Yang, Yifan and Huang, Qian and Qu, Ke},
  booktitle={2025 IEEE 41st International Conference on Data Engineering (ICDE)},
  pages={4455--4469},
  year={2025},
  organization={IEEE}
}

@inproceedings{han2025bridging,
  title={Bridging traffic state and trajectory for dynamic road network and trajectory representation learning},
  author={Han, Chengkai and Wang, Jingyuan and Wang, Yongyao and Yu, Xie and Lin, Hao and Li, Chao and Wu, Junjie},
  booktitle={Proceedings of the AAAI Conference on Artificial Intelligence},
  volume={39},
  number={11},
  pages={11763--11771},
  year={2025}
}

@inproceedings{licross,
  title={Cross City Traffic Flow Generation via Retrieval Augmented Diffusion Model},
  author={Li, Yudong and Wang, Jingyuan and Yu, Xie and Wang, Peiyu and Huang, Qian},
  booktitle={The Thirty-ninth Annual Conference on Neural Information Processing Systems},
  year={2025}
}

@inproceedings{yang2025hygmap,
  title={HygMap: Representing All Types of Map Entities via Heterogeneous Hypergraph},
  author={Yang, Yifan and Wang, Jingyuan and Yu, Xie and Tang, Yibang},
  booktitle={Proceedings of the Thirty-Fourth International Joint Conference on Artificial Intelligence},
  pages={9438--9446},
  year={2025}
}

@inproceedings{wang2025gtg,
  title={GTG: Generalizable Trajectory Generation Model for Urban Mobility},
  author={Wang, Jingyuan and Lin, Yujing and Li, Yudong},
  booktitle={Proceedings of the AAAI Conference on Artificial Intelligence},
  volume={39},
  number={1},
  pages={834--842},
  year={2025}
}

@inproceedings{jiang2023continuous,
  title={Continuous trajectory generation based on two-stage GAN},
  author={Jiang, Wenjun and Zhao, Wayne Xin and Wang, Jingyuan and Jiang, Jiawei},
  booktitle={Proceedings of the AAAI conference on artificial intelligence},
  volume={37},
  number={4},
  pages={4374--4382},
  year={2023}
}

@article{wang2021personalized,
  title={Personalized route recommendation with neural network enhanced search algorithm},
  author={Wang, Jingyuan and Wu, Ning and Zhao, Wayne Xin},
  journal={IEEE Transactions on Knowledge and Data Engineering},
  volume={34},
  number={12},
  pages={5910--5924},
  year={2021},
  publisher={IEEE}
}

@inproceedings{wang2019empowering,
  title={Empowering A* search algorithms with neural networks for personalized route recommendation},
  author={Wang, Jingyuan and Wu, Ning and Zhao, Wayne Xin and Peng, Fanzhang and Lin, Xin},
  booktitle={Proceedings of the 25th ACM SIGKDD international conference on knowledge discovery \& data mining},
  pages={539--547},
  year={2019}
}

@inproceedings{jiang2023self,
  title={Self-supervised trajectory representation learning with temporal regularities and travel semantics},
  author={Jiang, Jiawei and Pan, Dayan and Ren, Houxing and Jiang, Xiaohan and Li, Chao and Wang, Jingyuan},
  booktitle={2023 IEEE 39th international conference on data engineering (ICDE)},
  pages={843--855},
  year={2023},
  organization={IEEE}
}

@article{guo2020force,
  title={A force-directed approach to seeking route recommendation in ride-on-demand service using multi-source urban data},
  author={Guo, Suiming and Chen, Chao and Wang, Jingyuan and Ding, Yan and Liu, Yaxiao and Xu, Ke and Yu, Zhiwen and Zhang, Daqing},
  journal={IEEE Transactions on Mobile Computing},
  volume={21},
  number={6},
  pages={1909--1926},
  year={2020},
  publisher={IEEE}
}

@article{guo2023seeking,
  title={Seeking based on dynamic prices: Higher earnings and better strategies in ride-on-demand services},
  author={Guo, Suiming and Shen, Qianrong and Liu, Zhiquan and Chen, Chao and Chen, Chaoxiong and Wang, Jingyuan and Li, Zhetao and Xu, Ke},
  journal={IEEE Transactions on Intelligent Transportation Systems},
  volume={24},
  number={5},
  pages={5527--5542},
  year={2023},
  publisher={IEEE}
}

@article{wang2022shortening,
  title={Shortening passengers’ travel time: A dynamic metro train scheduling approach using deep reinforcement learning},
  author={Wang, Zhaoyuan and Pan, Zheyi and Chen, Shun and Ji, Shenggong and Yi, Xiuwen and Zhang, Junbo and Wang, Jingyuan and Gong, Zhiguo and Li, Tianrui and Zheng, Yu},
  journal={IEEE Transactions on Knowledge and Data Engineering},
  volume={35},
  number={5},
  pages={5282--5295},
  year={2022},
  publisher={IEEE}
}

@inproceedings{cheng2025poi,
  title={Poi-enhancer: An llm-based semantic enhancement framework for poi representation learning},
  author={Cheng, Jiawei and Wang, Jingyuan and Zhang, Yichuan and Ji, Jiahao and Zhu, Yuanshao and Zhang, Zhibo and Zhao, Xiangyu},
  booktitle={Proceedings of the AAAI conference on artificial intelligence},
  volume={39},
  number={11},
  pages={11509--11517},
  year={2025}
}

@inproceedings{wu2020learning,
  title={Learning effective road network representation with hierarchical graph neural networks},
  author={Wu, Ning and Zhao, Xin Wayne and Wang, Jingyuan and Pan, Dayan},
  booktitle={Proceedings of the 26th ACM SIGKDD international conference on knowledge discovery \& data mining},
  pages={6--14},
  year={2020}
}

@article{guo2019rod,
  title={Rod-revenue: Seeking strategies analysis and revenue prediction in ride-on-demand service using multi-source urban data},
  author={Guo, Suiming and Chen, Chao and Wang, Jingyuan and Liu, Yaxiao and Xu, Ke and Yu, Zhiwen and Zhang, Daqing and Chiu, Dah Ming},
  journal={IEEE Transactions on Mobile Computing},
  volume={19},
  number={9},
  pages={2202--2220},
  year={2019},
  publisher={IEEE}
}

@inproceedings{wu2019learning,
  title={Learning to effectively estimate the travel time for fastest route recommendation},
  author={Wu, Ning and Wang, Jingyuan and Zhao, Wayne Xin and Jin, Yang},
  booktitle={Proceedings of the 28th ACM International Conference on Information and Knowledge Management},
  pages={1923--1932},
  year={2019}
}

@article{zhang2024veccity,
  title={VecCity: A taxonomy-guided library for map entity representation learning},
  author={Zhang, Wentao and Wang, Jingyuan and Yang, Yifan and others},
  journal={arXiv preprint arXiv:2411.00874},
  year={2024}
}

\appendix

\section{Example Problem Instance}

\begin{figure}[t]
  \centering
  \includegraphics[page=1, width=\linewidth]{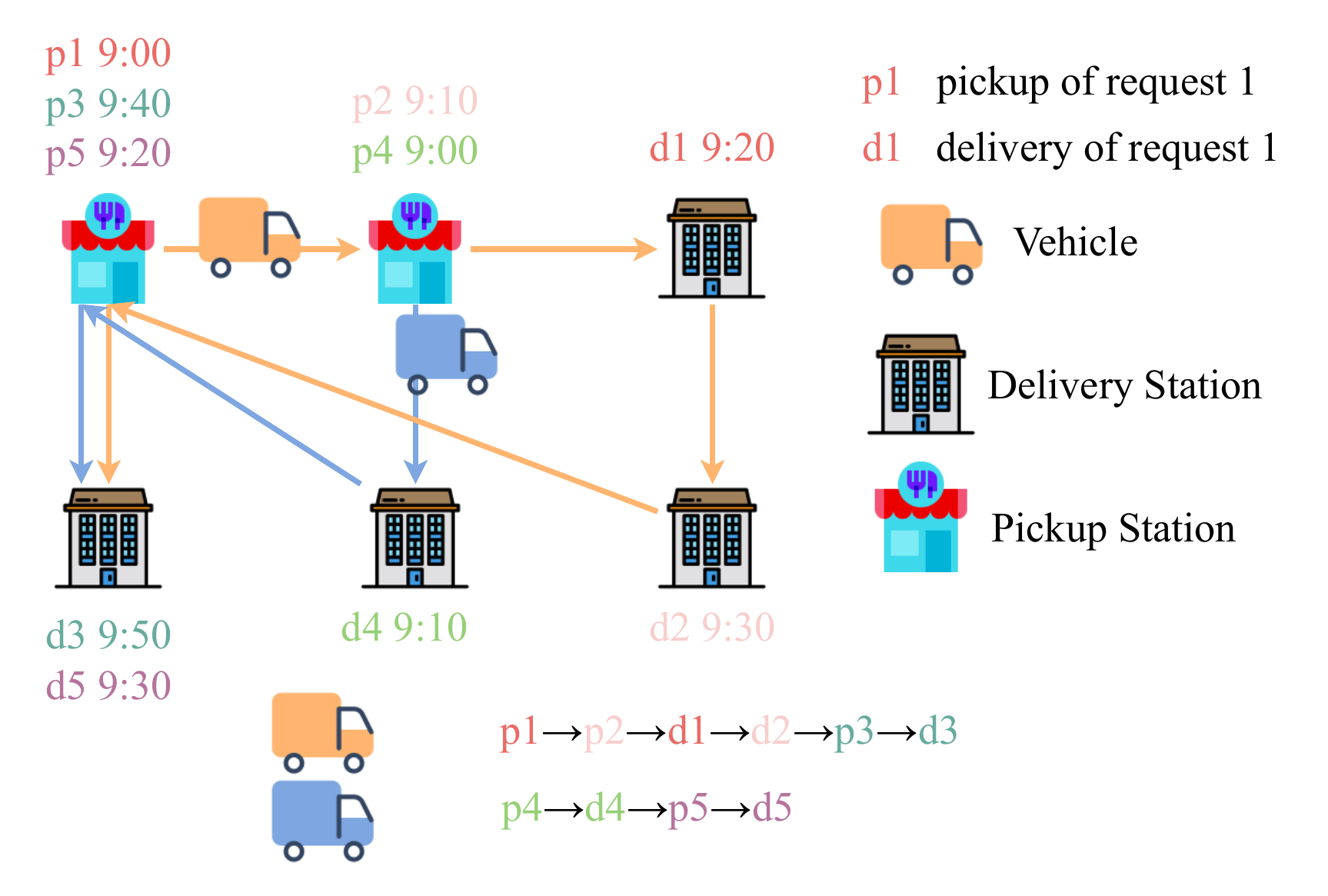}
  \caption{Schematic illustration of the MVDPDPSR problem.}
  \label{fig:scenario}
\end{figure}

Figure \ref{fig:scenario} presents a representative problem instance in our study. This instance consists of 2 pickup stations and 4 delivery stations, where each request is associated with one pickup station and one delivery station. It also includes 2 vehicles (yellow and blue) and 5 requests — each request is identified by a pair of pickup (\( p_n \)) and delivery (\( d_n \)) stations (with \( n = 1,2,3,4,5 \)), and has timestamps like \( 9{:}00 \), \( 9{:}10 \), etc.  
The yellow vehicle follows a sequential route: \( p_1 \rightarrow p_2 \rightarrow d_1 \rightarrow d_2 \rightarrow p_3 \rightarrow d_3 \) — as visualized by the orange arrows connecting these stations in the figure. It first handles early requests \( p_1 \) (pickup at \( 9{:}00 \)) and \( p_2 \) (pickup at \( 9{:}10 \)), then completes their deliveries (\( d_1 \) at \( 9{:}20 \), \( d_2 \) at \( 9{:}30 \)). When \( p_3 \) (pickup at \( 9{:}40 \)) dynamically arrives later, the yellow vehicle is scheduled to \textit{re-route back to the top-left pickup station} to fulfill this new request — highlighting how our problem requires adapting to \textbf{dynamically arriving requests}.  
Meanwhile, the blue vehicle executes another sequence: \( p_4 \rightarrow d_4 \rightarrow p_5 \rightarrow d_5 \) — as shown by the blue arrows. Here, \( p_4 \) (pickup at \( 9{:}00 \)) and \( d_4 \) (delivery at \( 9{:}10 \)) are processed first, followed by \( p_5 \) (pickup at \( 9{:}20 \)) and \( d_5 \) (delivery at \( 9{:}30 \)).  

When the blue vehicle reaches the upper-middle pickup station, Request 4 (e.g., \( p_4 \), the pickup for Request 4) is assigned to it, while Request 2 (e.g., \( p_2 \), the pickup for Request 2) is deferred and then assigned to the yellow vehicle upon its subsequent arrival — demonstrating the core challenge of \textbf{vehicle-request assignment}: we need to decide which vehicle each request should be assigned to.  

\section{Difference between Rolling Horizon and Markov Decision Process}

\begin{figure}[t]
  \centering
  \includegraphics[page=1, width=\linewidth]{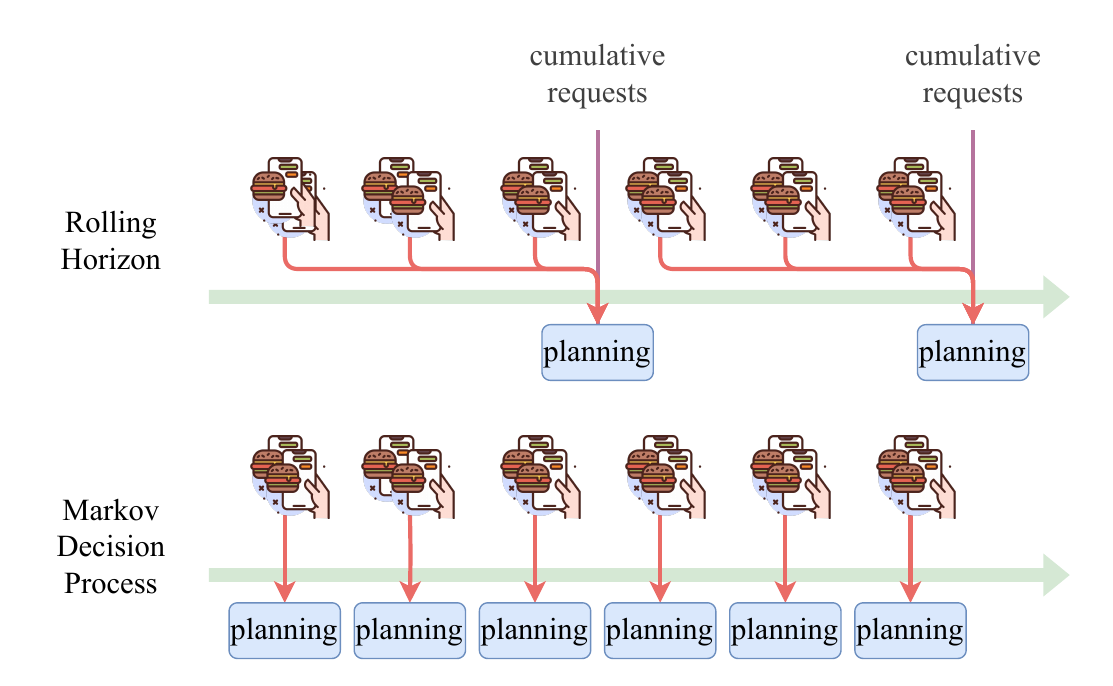}
  \caption{Comparison between Rolling Horizon and Markov Decision Process paradigms. The Rolling Horizon paradigm requires accumulating a sufficient number of requests before invoking a static solver for vehicle routing optimization, while the Markov Decision Process paradigm enables real-time decision-making as requests arrive.}
  \label{fig:decision}
\end{figure}

Rolling Horizon is often used in classical operations research methods. It transforms dynamic problems into static subproblems by accumulating a sufficient number of requests over a time window before planning begins (top of Fig.~\ref{fig:decision}).
During this accumulation already-arrived requests stay unprocessed causing unavoidable time waste. These static subproblems are then solved with exact algorithms or metaheuristics but their high computational complexity further hinders time efficiency. In contrast Markov Decision Process (MDP) triggers planning at each time step (bottom of Fig.~\ref{fig:decision}) and makes decisions right as requests arrive. The former fails to meet real-time dynamic decision-making needs due to its batch-based approach while the latter prioritizes real-time adaptation through frequent smaller-scale planning.

\section{Dataset Details}\label{appendix:dataset_details}
In the MVDPDPSR problem addressed in this paper, the problem scale is not solely determined by the number of stations but also depends on the number of requests, as we need to determine the assignment relationship between requests and vehicles. In the largest dataset, the number of requests reaches 800, which already exceeds the problem scale of most prior studies on static pickup-and-delivery problems, such as \cite{zong2022mapdp}.  
Below, we provide details of the synthetic datasets and the real-world datasets. Their statistics are summarized in Table \ref{tab:statistics}. 

\begin{itemize}[leftmargin=*]  
  \item {\em Synthetic Dataset}: We first synthesize some datasets to validate the effectiveness of the algorithm.  
  We generated $I \in \{20, 50, 300\}$ stations, with $I=20$, $50$, and $300$ corresponding to dataset suffixes \texttt{-S}, \texttt{-L}, and \texttt{-XL}, respectively. The distance between stations is sampled from $\mathrm{Uniform}\{0, 1, \dots, 10\}$ for \texttt{-S} datasets, from $\mathrm{Uniform}\{0, 1, \dots, 30\}$ for \texttt{-L} datasets and from $\mathrm{Uniform}\{0, 1, \dots, 20\}$ for \texttt{-XL} datasets. Then, the shortest path algorithm is applied.  
  The capacity of each vehicle was set to 3, and the number of vehicles was $K \in \{5, 15, 50\}$. Travel cost per unit distance is set to 0 or 0.3, with the latter case marked by an additional suffix \texttt{-cost}.  
  The origin and destination of each request were uniformly sampled from the $I$ stations, and the request's appearance time was sampled from $\mathrm{Uniform}\{1, 2, \dots, T\}$, where \( T \) is set to 58 for \texttt{-S} datasets and 128 for \texttt{-L} and \texttt{-XL} datasets. The profit for each request was the distance between its origin and destination.  

  \item {\em Real-World Dataset}: We used the DHRD\cite{assylbekov2023delivery} dataset, which consists of millions of food delivery requests from three cities: Taipei (\texttt{tpe}), Singapore (\texttt{sg}), and Stockholm (\texttt{se}). We treated each city as an independent dataset.  
  In this dataset, we treated each geographical area as a station, with areas divided by 5-character geohash granularity. After excluding stations with almost no requests, each dataset had $I = 36$ stations. The distance between two areas was defined as the minimum number of areas that needed to be traversed.  
  The capacity of each vehicle was set to 6, and the number of vehicles for the three cities was set to $K \in \{20, 15, 3\}$. In this dataset, we did not consider the $cost$ of vehicle travel, only the total profit of completed requests.  
  The origin and destination of each request were the stations corresponding to their respective geographical areas. We divided a day into $T=48$ time slots, corresponding to $T$ time steps in the simulation environment. The time step of a request's appearance was determined by its real-time slot. The profit for each request is set to 1.
  We divided the dataset into 76 days for training and validation, and 14 days for testing.  
\end{itemize}

\section{Computational Complexity Analysis}

The computational efficiency of MAPT is crucial for its practical application in dynamic vehicle routing scenarios, where real-time decision-making is essential. Our architecture maintains the fundamental structure of Transformer models while introducing minimal additional overhead through the Relation-Aware Attention and Informative Prior modules.
The encoder stage processes all entities (stations, requests, and vehicles) as a single sequence. For each encoder block with hidden dimension $H$, the complexity is dominated by the self-attention mechanism and feed-forward network operations. The self-attention computation scales as $O((I+M+K)^2H)$ where $I$, $M$, and $K$ represent the number of stations, requests, and vehicles respectively, while the feed-forward network contributes $O((I+M+K)H^2)$ per layer.

During decoding, we employ the KV-Cache technique commonly used in large language models to optimize autoregressive generation. This approach caches previously computed key-value pairs, significantly reducing redundant computations when processing sequential decisions. Each decoder block involves three main components: self-attention over the partially constructed solution (complexity $O((M+K)^2H)$), cross-attention between the decoder inputs and encoder outputs ($O((M+K)(I+M+K)H)$), and the feed-forward network operations ($O((M+K)H^2)$).

So the overall computational complexity of MAPT is $O(L[(I+M+K)^2H + (I+M+K)H^2])$, where $L$ represents the number of transformer layers. While this demonstrates quadratic scaling with respect to the number of entities, the parallel processing capabilities of modern GPUs enable efficient computation even for large problem instances.
This efficiency is particularly important for our target application, where the system must respond to dynamic requests in real-time while maintaining solution quality.

\begin{table}[t]
  \centering
  \tabcolsep=0.07cm
  \begin{tabular}{ccccccc}
    \toprule
    Scenio & Stations & Reqs. & Veh. & T & Cost & Profit \\
    \midrule
    synthetic-S & 20 & 110 & 5 & 58 & 0 & distance \\
    synthetic-S-cost & 20 & 110 & 5 & 58 & 0.3 & distance \\
    synthetic-L & 50 & 550 & 15 & 128 & 0 & distance \\
    synthetic-L-cost & 50 & 550 & 15 & 128 & 0.3 & distance \\
    synthetic-XL & 300 & 550 & 50 & 128 & 0 & distance \\
    \midrule
    dhrd-tpe & 36 & 800 & 20 & 48 & 0 & 1 \\
    dhrd-sg & 36 & 700 & 15 & 48 & 0 & 1 \\
    dhrd-se & 36 & 200 & 3 & 48 & 0 & 1 \\
    \bottomrule
  \end{tabular}
  \caption{Statistics of experimental datasets. \textit{cost} is the cost per unit distance; \textit{profit} is the profit per request (either distance-based or fixed).}
  \label{tab:statistics}
\end{table}

\section{Training Algorithm} \label{appendix:training_algorithm}

The main part of our training algorithm follows the original Proximal Policy Optimization (PPO) algorithm \cite{schulman2017proximal}, and we have incorporated some learning rate schedule strategies. We summarize our training process in Algorithm \ref{alg:ppo_training}.

\begin{algorithm}[t]
\caption{PPO Training Algorithm for MAPT}
\label{alg:ppo_training}
\begin{algorithmic}[1]
\STATE \textbf{Initialize:} 
\STATE \quad Policy network $\pi_{\theta}$ with parameters $\theta$.
\STATE \quad Replay buffer $\mathcal{B}$.
\STATE \quad Environment $\mathcal{E}$, episode length $T$.
\STATE \quad Learning rate scheduler with warmup and linear decay.
\STATE \quad Rollout iterations $K$, rollout threads $N$.

\FOR{iteration $k = 1, 2, \dots, K$}
    \FOR{thread $n = 1, 2, \dots, N$}
        \STATE Reset environment $\mathcal{E}$ and initialize observations $s_0$.
        \FOR{step $t = 0, 1, 2, \dots, T-1$}
            \STATE \textbf{Collect Data:}
            \STATE \quad Sample actions $a_t$ from policy $\pi_{\theta}(a_t|s_t)$.
            \STATE \quad Execute actions $a_t$ in environment $\mathcal{E}$.
            \STATE \quad Observe next state $s_{t+1}$ and reward $r_t$.
            \STATE \quad Store transition $(s_t, a_t, r_t, s_{t+1})$ in buffer $\mathcal{B}$.
        \ENDFOR
    \ENDFOR

    \STATE \textbf{Compute Advantages:}
    \STATE \quad Compute advantages using Generalized Advantage Estimation (\ref{eq:gae}).

    \FOR{epoch $e = 1, 2, \dots, E$}
        \STATE \textbf{Optimize Surrogate Objective:}
        \STATE \quad Sample mini-batch of transitions from buffer $\mathcal{B}$.
        \STATE \quad Compute critic loss by (\ref{eq:critic_loss}) and policy loss by (\ref{eq:actor_loss}).
        \STATE \quad Update policy $\pi_{\theta}$ using gradient descent on joint loss (\ref{eq:total_loss}).
    \ENDFOR

    \STATE \textbf{Update Old Policy:}
    \STATE \quad $\theta_{\text{old}} \leftarrow \theta$
    \STATE \quad update learning rate through scheduler.
\ENDFOR
\end{algorithmic}
\end{algorithm}

\section{Implementation Details}

For the model architecture, we employ a Transformer with 6 encoder layers and 2 decoder layers. The hidden size is set to 128, and the number of attention heads is 2. 
Regarding the optimizer configuration, we utilize the Adam optimizer with a learning rate schedule that includes a warm-up phase and a decay phase. Specifically, the learning rate increases linearly during the first quarter of the total training steps, reaching the target learning rate, and subsequently undergoes linear decay for the remaining steps.
For the Proximal Policy Optimization (PPO) algorithm, we set the discount factor $\gamma = 0.99$, the clipping coefficient $\epsilon = 0.2$, and the Generalized Advantage Estimation (GAE) parameter $\lambda = 0.99$. Critic loss coefficient $\alpha$ is set to 1. The number of PPO epochs is fixed at 1.
The hyperparameter $\beta$ is set to 0.03. During training, we monitor the model's performance on the validation set and retain the best-performing model as the final model. Additional dataset-specific hyperparameters are provided in Table \ref{tab:dataset_hyperparam}.
All experiments are conducted using Python 3.10 and PyTorch 2.5.

\begin{table}[t]
  \centering
  \tabcolsep=3pt
  \begin{tabular}{lcccc}
    \toprule
    Dataset & Total & Rollout & MiniBatch & LR \\
    \midrule
    synth-S & 16K & 256 & 256 & 1e-4 \\
    synth-S-cost & 16K & 256 & 256 & 1e-4 \\
    synth-L & 4K & 64 & 64 & 2e-5 \\
    synth-L-cost & 4K & 64 & 64 & 5e-5 \\
    synth-XL & 4K & 64 & 64 & 2e-5 \\
    dhrd-se & 8K & 256 & 256 & 2e-5 \\
    dhrd-tpe & 2K & 64 & 64 & 2e-5 \\
    dhrd-sg & 2K & 64 & 64 & 1e-5 \\
    \bottomrule
  \end{tabular}
  \caption{Dataset Specific Hyper-parameters. \textit{Total} indicates the overall training budget. \textit{Rollout} specifies the experience collection per iteration. \textit{Minibatch} determines the SGD batch size. \textit{LR} controls the policy update step size.}
  \label{tab:dataset_hyperparam}
\end{table}

\section{Difference between Non-AutoRegressive and AutoRegressive Decoding} \label{appendix:compare_ar}
\begin{figure*}[t]
  \centering
  \includegraphics[page=1, width=0.7\textwidth]{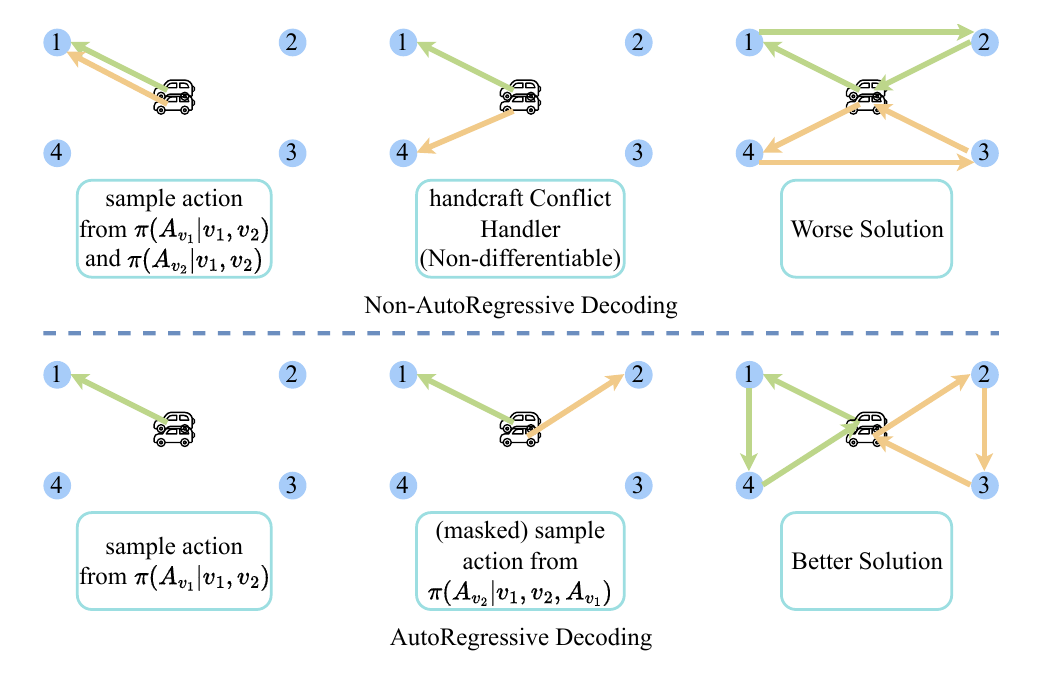}
  \caption{Comparison between Non-AutoRegressive Decoding(up) and AutoRegressive Decoding(down).}
  \label{fig:compare_ar}
\end{figure*}

We use a simple VRP example to illustrate the difference between the Non-AutoRegressive method with independent decoding and our AutoRegressive method in terms of decoding, as shown in Figure \ref{fig:compare_ar}.
Suppose two vehicles, \(v_1\) and \(v_2\), are at the same location and need to collaboratively visit stations \(n_1 \dots n_4\).
The optimal solution is achieved when both vehicles move left and right respectively, while a suboptimal solution is obtained when they move up and down respectively.
Let the actions of the two vehicles be \(A_{v_1}\) and \(A_{v_2}\), representing the next station they move to at the first step, and \(\pi\) represent a scheduling policy.
However, for the Non-AutoRegressive independent decoding method, it can only model the probability distribution of each vehicle's individual actions \(\pi(A_{v_1} \mid v_1, v_2)\) and \(\pi(A_{v_2} \mid v_1, v_2)\), but fails to model the joint probability distribution of the two vehicles' actions \(\pi(A_{v_1}, A_{v_2} \mid v_1, v_2)\).
Consequently, it cannot guarantee that both vehicles move left and right respectively. Moreover, since a conflict handler is applied afterward, its gradient cannot be backpropagated, making it impossible to optimize the neural network parameters using conflict information when conflicts occur, which limits the capability of such methods.
For AutoRegressive decoding, we first decode the action \(A_{v_1}\) of \(v_1\) according to \(\pi(A_{v_1}=n_1\dots n_4 \mid v_1, v_2)\), then use \(A_{v_1}\) as an additional condition to decode the action \(A_{v_2}\) of \(v_2\), i.e., \(\pi(A_{v_2}=n_1\dots n_4 \mid v_1, v_2, A_{v_1})\).
At this point, the policy can determine the direction of \(v_2\) based on whether \(A_{v_1}\) is left or right, thus achieving the optimal solution. Additionally, we can mask out \(A_{v_1}\) when decoding \(A_{v_2}\) to avoid conflicts.

\section{Does MAPT Overly Rely on Informative Prior?}
\begin{figure}[t]
\centering  
\begin{subfigure}[b]{0.49\linewidth}
  \includegraphics[width=\textwidth]{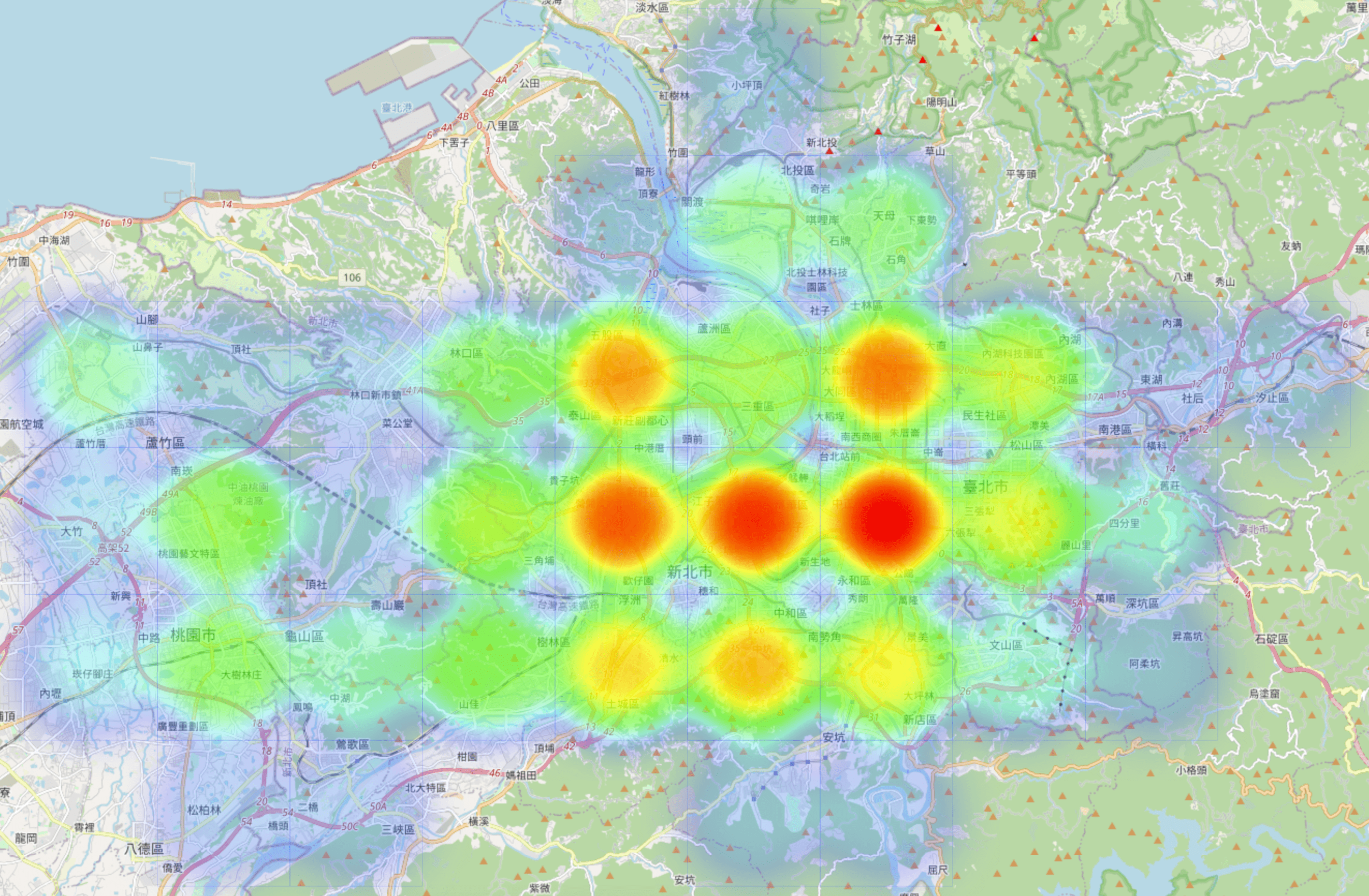}
  \caption{}
\end{subfigure}
\begin{subfigure}[b]{0.49\linewidth}
  \includegraphics[width=\textwidth]{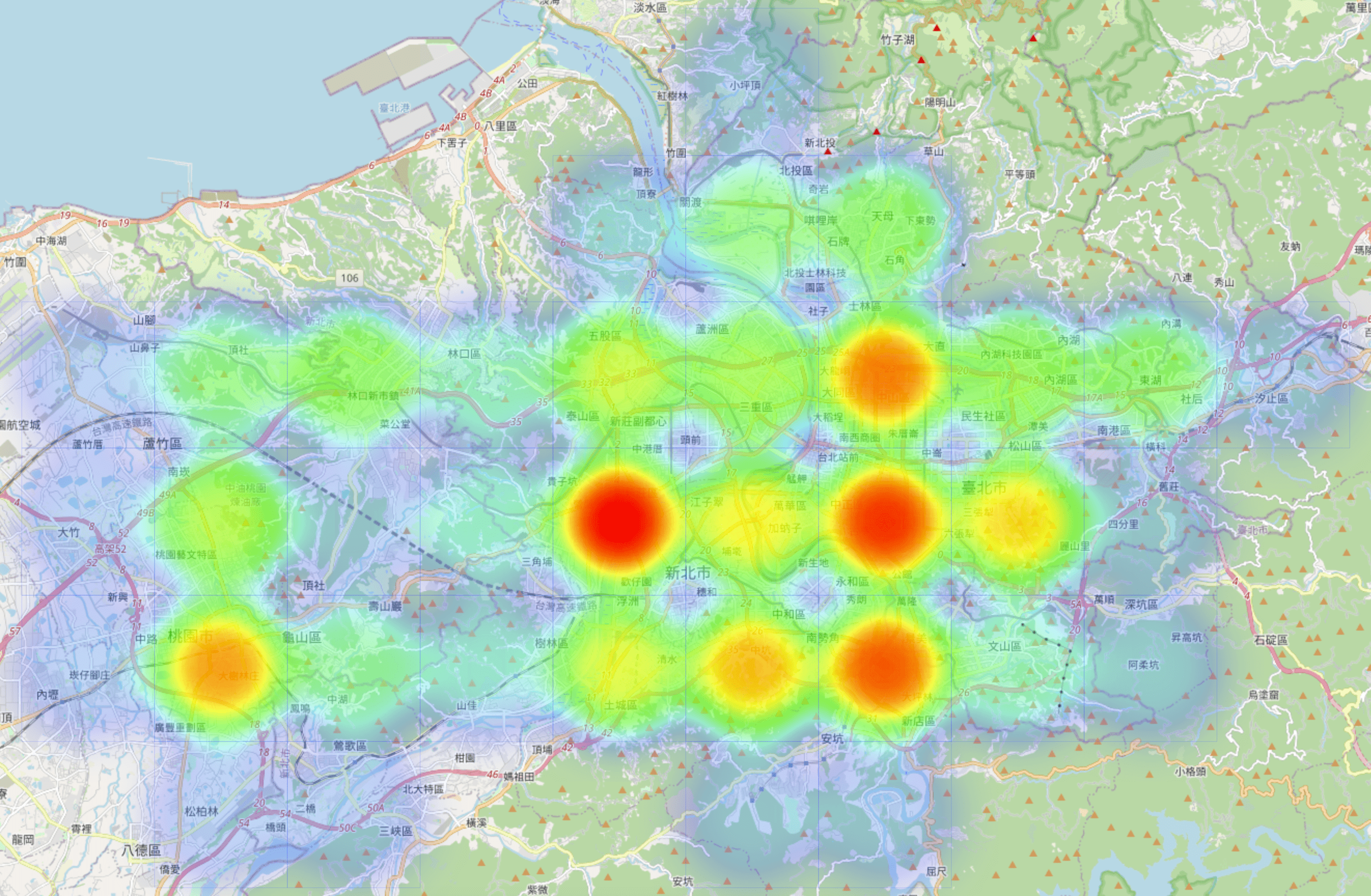}
  \caption{}
\end{subfigure}
\caption{Heatmap of the distribution of request origins(a) and destinations(b) in the dhrd-tpe dataset, where darker colors indicate higher quantities.}
\label{fig:ori_des}
\end{figure}

To verify whether MAPT excessively relies on the Informative Prior, we compare the experimental results between MAPT and using only the Informative Prior as the decision-making algorithm, as shown in Table \ref{tab:compare_heur}.
In synthetic datasets, problem instances are randomly generated, making it difficult to form complex and realistic distributions. As a result, the generated problem instances tend to be simpler, leading to cases where the difference between the Informative Prior and MAPT is small.
However, for real-world datasets, our model (MAPT) shows significant improvements over the Informative Prior, with enhancement rates of 14\% for dhrd-tpe, 16\% for dhrd-sg, and 19\% for dhrd-se.
In Figure \ref{fig:ori_des}, we show the distribution of request origins and destinations in the dhrd-tpe dataset, which differs significantly from the Uniform distribution of synthetic data. 
Reinforcement learning can learn more complex distributions based on the Informative Prior, which is why its improvement on real-world datasets is more significant than on synthetic datasets. 
Moreover, even for the large-scale synthetic-XL dataset, our model achieves a notable 37\% improvement over the Informative Prior.  
These results demonstrate that our model does not overly rely on Informative Priors.
\begin{table*}[t]
  \tabcolsep=0.02cm
  \centering
{
  \fontsize{9pt}{10.8pt}\selectfont
  \begin{tabular}{c|cc|cc|cc|cc|cc|cc|cc|cc}
    \toprule
    Scenario & \multicolumn{2}{c|}{synth-S} & \multicolumn{2}{c|}{synth-S-cost} & \multicolumn{2}{c|}{synth-L} & \multicolumn{2}{c|}{synth-L-cost} & \multicolumn{2}{c|}{dhrd-tpe} & \multicolumn{2}{c|}{dhrd-sg} & \multicolumn{2}{c|}{dhrd-se} & \multicolumn{2}{c}{synth-XL} \\
    Metric & Obj. $\uparrow$ & Comp. $\uparrow$ & Obj. $\uparrow$ & Comp. $\uparrow$ & Obj. $\uparrow$ & Comp. $\uparrow$ & Obj. $\uparrow$ & Comp. $\uparrow$ & Obj. $\uparrow$ & Comp. $\uparrow$ & Obj. $\uparrow$ & Comp. $\uparrow$ & Obj. $\uparrow$ & Comp. $\uparrow$ & Obj. $\uparrow$ & Comp. $\uparrow$ \\
\midrule
Infor Prior & 267.1 & 0.81 & 185.0 & 0.81 & 1858.9 & 0.79 & 1305.5 & 0.79 & 614.1 & 0.77 & 323.7 & 0.62 & 66.1 & 0.54 & 2340.7 & 0.47 \\
MAPT & \textbf{275.2} & \textbf{0.83} & \textbf{192.1} & \textbf{0.84} & \textbf{1875.1} & \textbf{0.80} & \textbf{1348.6} & \textbf{0.81} & \textbf{697.2} & \textbf{0.87} & \textbf{376.1} & \textbf{0.71} & \textbf{78.9} & \textbf{0.64} & \textbf{3227.5} & \textbf{0.65} \\

\bottomrule
\end{tabular}
\caption{Performance comparison between Informative Prior and MAPT.}
\label{tab:compare_heur}
}
\end{table*}

\section{Generalization}

\begin{table}[t]
  \centering
  \tabcolsep=0.04cm
  \begin{tabular}{c|cc|cc|cc}
    \toprule
    Scenario & \multicolumn{2}{c|}{synth-S} & \multicolumn{2}{c|}{dhrd-se} & \multicolumn{2}{c}{synth-XL} \\
    Metric & Obj. $\uparrow$ & Comp. $\uparrow$ & Obj. $\uparrow$ & Comp. $\uparrow$ & Obj. $\uparrow$ & Comp. $\uparrow$ \\
    \midrule
    synth-S & \textbf{275.2} & \textbf{0.83} & 74.4 & 0.61 & 2470.2 & 0.50 \\
    dhrd-se & 117.4 & 0.36 & \textbf{78.9} & \textbf{0.64} & 1480.0 & 0.30 \\
    synth-XL & 109.8 & 0.34 & 52.6 & 0.42 & \textbf{3227.5} & \textbf{0.65} \\
  \bottomrule
\end{tabular}
\caption{Generalization of MAPT. The row index indicates that the model is trained on the corresponding dataset, and the column index indicates that the model is evaluated on the corresponding dataset.}
\label{tab:transfer}  
\end{table}

Since our decision-making does not strictly depend on the number of vehicles and requests, we expect our model to have some generalization ability. To this end, we conducted a series of experiments:
\begin{itemize}[leftmargin=*]
  \item \textbf{synthetic-S $\rightarrow$ synthetic-XL} represents the validation of whether a model trained on small-scale data can be used on large-scale problems.
  \item \textbf{synthetic-S $\rightarrow$ dhrd-se} represents the validation of whether a model trained on data from one distribution can generalize to other distributions.
\end{itemize}
The generalization results are shown in Table \ref{tab:transfer}. The experiments show that models trained on small synthetic dataset can generalize to real-world dataset and large-scale dataset to some extent.
This is due to two reasons: first, the use of Transformer, a model structure with good generalization capabilities, and second, the inclusion of informative priors in the model, which provides guidance in unseen distributions.
It is worth noting that when tested on real-world datasets and large-scale datasets, our method still outperforms the pure Informative Priors approach (see Tables \ref{tab:compare_heur} and \ref{tab:transfer}), proving that our deep model has also learned some transferable knowledge, rather than relying solely on the Informative Priors for generalization.
Although the model trained on the dhrd-se dataset performed best on this dataset, it performed poorly when generalized to synthetic datasets. We speculate that this is because the request OD distribution and station embeddings learned from the dhrd-se dataset are not suitable for synthetic datasets.
However, there is still much to explore in the field of generalization as part of our future research.

\begin{table}[t]
  \centering
  \tabcolsep=3pt
  \begin{tabular}{lcc}
    \toprule
    Dataset & Horizon & Replan\\
    \midrule
    synth-S & 20 &  10 \\
    synth-S-cost & 20 & 10 \\
    synth-L & 60 & 30 \\
    synth-L-cost & 60 & 30 \\
    synth-XL & 40 & 20 \\
    dhrd-se & 30 & 15 \\
    dhrd-tpe & 30 & 15 \\
    dhrd-sg & 30 & 15 \\
    \bottomrule
  \end{tabular}
  \caption{Rolling Horizon Hyper-parameters. \textit{Horizon} indicates the maximum allowable planning horizon. \textit{Replan} specifies the re-planning interval.}
  \label{tab:rh_hyperparam}
\end{table}

\section{Baselines Implementation}

\subsubsection{Rolling Horizon}
The Rolling Horizon Policy addresses the dynamic nature of the Multi-Vehicle Dynamic Pickup and Delivery Problem by breaking it into smaller, static subproblems within a sliding time window, enabling the use of traditional optimization methods.
At each decision point, the policy determines a planning horizon based on the remaining time frames and the maximum allowable planning horizon, ensuring computational tractability. The maximum allowable planning horizon used for each dataset is shown in Table \ref{tab:rh_hyperparam}, column \textit{Horizon}.
The problem within this window is formulated as a static instance, where vehicles, requests, and constraints (e.g., vehicle capacity, request appearance times, and preloaded requests) are fixed for the duration of the window.
A solver, chosen from a set of available methods (e.g., \textbf{OR-Tools}, \textbf{Genetic Algorithm(GA)}, or \textbf{Simulated Annealing(SA)}), is then applied to optimize the assignment of requests to vehicles and their routes within the window.
The solver’s solution is translated into actionable decisions, such as assigning requests to vehicles and determining the next hop for each vehicle, while ensuring feasibility with respect to constraints like vehicle capacity and request uniqueness.
The policy periodically re-evaluates and re-plans at predefined intervals to adapt to new information and changing conditions. The re-planning interval used for each dataset is shown in Table \ref{tab:rh_hyperparam}, column \textit{Replan}.
By iteratively solving these static subproblems and updating the solution as the horizon shifts, the Rolling Horizon Policy effectively handles the dynamic and stochastic nature of the problem, balancing computational efficiency with solution quality.

\subsubsection{OR-Tools}
We address the Multi-Vehicle Dynamic Pickup and Delivery Problem with Stochastic Requests using the Google OR-Tools CP-SAT solver.
The problem involves a fleet of vehicles \( K \), a set of requests \( M \), and a set of stations \( N \) over a time horizon \( T \).
Each vehicle \( k \in K \) has a starting location \( start_k \), a capacity \( capacity_k \), and a join time \( jointime_k \) indicating when it becomes available for the first decision-making.
Requests \( m \in M \) appear at time \( appear_m \) and must be picked up from \( from_m \) and delivered to \( to_m \).
Some requests may be pre-loaded on vehicles, denoted by \( preload_{k,m} \), and vehicles are en route until their join time, during which they cannot perform any actions.
For the convenience of modeling, we assume that the volume of each request is 1.

Decision Variables:
\begin{itemize}[leftmargin=*]
    \item \( pickup_{k,m,t} \): Binary variable indicating whether vehicle \( k \) picks up request \( m \) at time \( t \).
    \item \( delivery_{k,m,t} \): Binary variable indicating whether vehicle \( k \) delivers request \( m \) at time \( t \).
    \item \( load_{k,t} \): Integer variable representing the load of vehicle \( k \) at time \( t \).
    \item \( loc_{k,t} \): Integer variable representing the location of vehicle \( k \) at time \( t \).
    \item \( noact_{k,t} \): (No Action) Binary variable indicating whether vehicle \( k \) is not performing any action at time \( t \), e.g. the vehicle is en route.
\end{itemize}

Constraints:
\begin{itemize}[leftmargin=*]
    \item \textbf{Pre-Loaded Requests}: For pre-loaded requests, the pickup action is enforced at \( t = -1 \):
    \begin{gather}
        pickup_{k,m,-1} = preload_{k,m} \quad \forall k \in K, m \in M \\
        load_{k, -1} = \sum_{m} preload_{k, m} \quad \forall k \in K
    \end{gather}

    \item \textbf{Load Dynamics}: The load of each vehicle is updated based on pickup and delivery actions:
    \begin{equation}
      \begin{split}
          load_{k,t} = load_{k,t-1} + \sum_{m \in M} pickup_{k,m,t} - \\ \sum_{m \in M} delivery_{k,m,t} \forall k \in K, t \in [0, T]
      \end{split}
    \end{equation}

    \item \textbf{Location Dynamics}: The location of each vehicle is updated based on pickup, delivery, and no-action variables:
    \begin{equation}
      \begin{split}
          loc_{k,t} = 
          \begin{cases}
              start_k & \text{if } t \le jointime_k \\
              loc_{k,t-1} & \text{if } noact_{k,t-1} = 1 \\
              to_m & \text{if } delivery_{k,m,t} = 1 \\
              from_m & \text{if } pickup_{k,m,t} = 1
          \end{cases} \\
          \forall k \in K, t \in [0, T], m \in M
      \end{split}
    \end{equation}

    \item \textbf{No-Action Constraints}: If no pickup or delivery occurs at time \( t \), the no-action variable is set to 1:
    \begin{equation}
      \begin{split}
          noact_{k,t} = 1 \iff \\
          \sum_{m \in M} pickup_{k,m,t} + \sum_{m \in M} delivery_{k,m,t} = 0 \\
          \forall k \in K, t \in [0, T]
      \end{split}
    \end{equation}

    \item \textbf{Request Assignment}: Each request must be picked up and delivered by at most one vehicle:
    \begin{equation}
        \sum_{k \in K} \sum_{t \in [-1, T]} pickup_{k,m,t} \leq 1 \quad \forall m \in M
    \end{equation}
    \begin{equation}
        \sum_{k \in K} \sum_{t \in [-1, T]} delivery_{k,m,t} \leq 1 \quad \forall m \in M
    \end{equation}

    \item \textbf{Pickup-Delivery Sequence}: A request must be picked up before it can be delivered:
    \begin{equation}
      \begin{split}
          \sum_{t' \in [-1, t]} pickup_{k,m,t'} = 1 \implies delivery_{k,m,t} = 1 \\
          \forall k \in K, m \in M, t \in [-1, T]
      \end{split}
    \end{equation}

    \item \textbf{Capacity Constraints}: The load of each vehicle must not exceed its capacity:
    \begin{equation}
        load_{k,t} \leq capacity_k \quad \forall k \in K, t \in [0, T]
    \end{equation}

    \item \textbf{Time-Distance Constraints}: The time between consecutive actions must be sufficient to travel between locations:
    \begin{equation}
      \begin{split}
          t_1 - t_0 \geq dist_{loc_{k,t_0}, loc_{k,t_1}} \\ \forall k \in K, t_0, t_1 \in [0, T], t_0 < t_1
      \end{split}
    \end{equation}
    
    \item \textbf{Join Time Constraints}: Vehicles cannot perform any actions before their join time:
    \begin{equation}
        noact_{k,t} = 1 \quad \forall k \in K, t \in [0, jointime_k - 1]
    \end{equation}
\end{itemize}

Objective Function:
The objective is to maximize the total profit of delivered requests while minimizing travel costs:
\begin{equation}
  \begin{split}
      \text{Maximize } \sum_{m \in M} \sum_{k \in K} \sum_{t \in [0, T]} delivery_{k,m,t} \cdot value_m \\ - \sum_{k \in K} \sum_{t \in [0, T-1]} cost_{loc_{k,t}, loc_{k,t+1}} \\
  \end{split}
\end{equation}

\subsubsection{Simulated Annealing}
The Simulated Annealing (SA) algorithm for solving the Multi-Vehicle Dynamic Pickup and Delivery Problem with Stochastic Requests starts by generating an initial solution through random request assignments to vehicles, ensuring constraints such as initial positions, displacement, location consistency, request uniqueness, capacity, request appearance times, and vehicle start times are satisfied.
To construct a neighbor solution, the algorithm randomly selects a request and removes its current assignment from the vehicle’s path.
Then the selected request is reassigned to a new vehicle, and both pickup and delivery times are randomly selected within their respective feasible windows, ensuring the pickup occurs after the request’s appearance time and the delivery occurs after the pickup plus the travel time.
The annealing process iteratively explores the solution space by accepting better solutions or probabilistically accepting worse solutions based on the current temperature, which gradually decreases according to a predefined cooling rate.
The initial temperature is set to 1000, the final temperature is set to 1, the cooling rate is set to 0.99, and the maximum number of iterations is set to 5000.

\subsubsection{Genetic Algorithm}
The Genetic Algorithm (GA) for solving the Multi-Vehicle Dynamic Pickup and Delivery Problem with Stochastic Requests begins by generating an initial population of feasible solutions, where requests are randomly assigned to vehicles while ensuring constraints such as initial positions, displacement, location consistency, request uniqueness, capacity, request appearance times, and vehicle start times are satisfied.
Each individual in the population represents a solution, and its fitness is evaluated using an exponential transformation of the objective function, which combines the total profit of served requests and the cost of vehicle movements. The algorithm evolves the population through selection, crossover, and mutation operations. During selection, individuals are chosen probabilistically based on their fitness, favoring better solutions.
The crossover operation randomly selects a request and swaps its assignment between two parent solutions, ensuring feasibility by updating vehicle locations and loads.
The mutation operation randomly selects a request and reallocates it to a different vehicle or adjusts its pickup and delivery times, maintaining feasibility through constraint checks.
The population iteratively evolves over a predefined number of generations, with the best solution being tracked and updated based on fitness.
The population size is set to 10, and the number of generations is set to 500.

\end{document}